%% file: am-parsing-20.tex
\documentclass[11pt,a4paper]{article}
\usepackage[hyperref]{emnlp2020}
\usepackage{times}
\usepackage{latexsym}

\usepackage{microtype}

\aclfinalcopy 
\def\aclpaperid{2419} 

\include{macros}

\title{Fast semantic parsing with well-typedness guarantees}

\author{Matthias Lindemann \and Jonas Groschwitz \and Alexander Koller\\
  Department of Language Science and Technology\\
  Saarland University\\
  \url{{mlinde|jonasg|koller}@coli.uni-saarland.de}}

\date{}

\begin{document}
\maketitle

\begin{abstract}
  AM dependency parsing is a linguistically principled method for neural semantic parsing
  with high accuracy
  across multiple graphbanks. 
  It relies 
  on a type system that~models semantic valency but makes existing parsers slow.
  We describe an A* parser and a transi\-tion-based
  parser for AM dependency par\-sing which guarantee well-typedness and 
  improve parsing speed by up to 3 orders of magnitude, while maintaining or improving accuracy.
\end{abstract}

\input introduction
\input relwork

\input background

\input astar
\input topdown

\input evaluation

\input conclusion

\paragraph{Acknowledgments.}
We thank the anonymous reviewers and the participants of the DELPH-IN
Summit 2020 for their helpful feedback and comments. We thank Rezka
Leonandya for his work on an earlier version of the A*
parser. This research was funded by the Deutsche
Forschungsgemeinschaft (DFG, German Research Foundation), project
KO~2916/2-2.

\pagebreak
\bibliographystyle{acl_natbib}
\bibliography{mybib,emnlp2020}

\newpage
\appendix
\input appendix-results

\input appendix-hyperparams

\input appendix-am
\input proofs-new

\end{document}

%% file: macros.tex
\usepackage{microtype}

\usepackage{qtree}
\usepackage{forest}

\usepackage{float}
\usepackage{graphicx} 
\usepackage{amsmath}
\usepackage{amsthm}
\usepackage{amssymb}
\usepackage{latexsym}
\usepackage{mathtools}

\usepackage{algpseudocode}
\usepackage{algorithm}

\usepackage{booktabs}

\usepackage{caption}
\usepackage{subcaption}

\usepackage{bussproofs} 
\usepackage{stackengine}

\usepackage{enumerate}
\usepackage{multirow}

\newcommand{\ie}{i.e.~}

\newcommand{\sortof}[1]{`#1'}
\newcommand{\of}[1]{\left(#1\right)}

\newlength\mylen

\makeatletter
\def\th@definition{%
  \thm@notefont{}
  \normalfont 
}
\makeatother

\theoremstyle{definition} 

\newtheorem{thm}{Theorem}[section]
\newtheorem{lemma}[thm]{Lemma}

\newtheorem{assumption}{Assumption}

\theoremstyle{definition}

\usepackage{amssymb}

\usepackage{pifont}

\usepackage[pdf,singlefile]{graphviz}

\usepackage{setspace}

\newcommand{\concat}{\ensuremath{\oplus}}

\usepackage{bm}
\usepackage{xspace}
\usepackage[utf8]{inputenc}

\newcommand{\tp}[1]{\ensuremath{\tau_{#1}}}   
\newcommand{\type}[1]{\ensuremath{[#1]}}

\newcommand{\app}[1]{\text{\textsc{App}}\ensuremath{_{#1}}}  
\newcommand{\modify}[1]{\text{\textsc{mod}}\ensuremath{_{#1}}\xspace}
\newcommand{\lignore}{\text{\textsc{ignore}}}
\newcommand{\lroot}{\textsc{root}}

\newcommand{\src}[1]{\text{\textsc{#1}}\xspace}                                  

\newcommand{\G}[1]{\ensuremath{G_{\text{#1}}}} 

\newcommand{\emptytype}{\ensuremath{[\,]}}

\newcommand{\el}[1]{\text{`#1'}}

 \newcommand{\Span}[2]{[#1,#2]}
 \newcommand{\oper}{\ensuremath{\ell}}  
\newcommand{\edge}[3]{\ensuremath{#1\xrightarrow{\text{\tiny #3}}#2}}
\newcommand{\Cons}{\ensuremath{C}}
\newcommand{\Sources}{\ensuremath{S}}
\newcommand{\typeSet}{\ensuremath{{\Omega}}}
\newcommand{\edgeLabels}{\ensuremath{L}}

\newcommand{\oneover}[2]{\ensuremath{\displaystyle \begin{array}{@{}c@{}} #1 \\
 		\hline #2  \end{array}}}
\newcommand{\infer}[3]{\oneover{#2}{#3} \;\;{\mbox{#1}}}

\newcommand{\pitem}[5]{(\Span{#1}{#2}, #3, #4):#5}

\newcommand{\cost}{\ensuremath{c}}
\newcommand{\edgeCost}[3]{\ensuremath{\cost\of{\edge{#1}{#2}{#3}}}} 
\newcommand{\graphCost}[2]{\ensuremath{\cost\of{#2,#1}}} 

\usepackage{tikz}
\usepackage{tikz-qtree}

\usetikzlibrary{graphs}
\usetikzlibrary{shapes,arrows}
\usetikzlibrary{positioning}
\usetikzlibrary{quotes}

\tikzset{snode/.style={
    ellipse,
    minimum size=6mm,
    very thick,
    draw=black,
    font=\rmfamily}}

\tikzset{ssrc/.style={
    ellipse,
    minimum size=6mm,
    very thick,
    fill=black!10,
    draw=black,
    text=black,
    font=\rmfamily}}

\tikzset{sanno/.style={node distance=-1mm,font=\rmfamily\small}}

\tikzset{sedgel/.style={
    color=black,
    sloped, above,
    font=\rmfamily\small}}

\tikzset{sedge/.style={
    thick,
    >=stealth'
}}

\usepackage{array}

\newcommand{\cin}[1]{\scalebox{0.9}{\textcolor{gray}{\ensuremath{\pm #1}}}}

\usepackage{color, colortbl}
\usepackage{bm}
\definecolor{orange}{rgb}{1,0.5,0}
\definecolor{mdgreen}{rgb}{0.05,0.6,0.05}
\definecolor{mdblue}{rgb}{0,0,0.7}
\definecolor{dkblue}{rgb}{0,0,0.5}
\definecolor{dkgray}{rgb}{0.3,0.3,0.3}
\definecolor{slate}{rgb}{0.25,0.25,0.4}
\definecolor{gray}{rgb}{0.5,0.5,0.5}
\definecolor{ltgray}{rgb}{0.7,0.7,0.7}
\definecolor{purple}{rgb}{0.7,0,1.0}
\definecolor{lavender}{rgb}{0.65,0.55,1.0}
\definecolor{brown}{rgb}{0.6,0.2,0.2}

\newcommand{\ignore}[1]{}

\newcommand{\lextypO}[1]{\ensuremath{\textcolor{red}{#1}}}
\newcommand{\ttypO}[1]{\ensuremath{\textcolor{blue}{#1}}}

\newcommand{\greek}[1]{\begingroup\setuplatintogreek#1\endgroup}

\newcommand{\setuplatintogreek}{%
  \mathcode`t=\tau
  \mathcode`l=\lambda
}
\newcommand{\lextyp}[1]{\lextypO{\greek{#1}}}
\newcommand{\ttyp}[1]{\ttypO{\greek{#1}}}

\newcommand{\applyset}{\ensuremath{\mathcal{A}}}

\newcommand{\ledge}[3]{\ensuremath{#1 \xrightarrow{#2} #3}}
\newcommand{\req}[2]{\ensuremath{\text{\textit{req}}_{#2}(#1)}}

\newcommand{\pE}{\ensuremath{\mathbb{E}}}
\newcommand{\pT}{\ensuremath{\mathbb{T}}}
\newcommand{\pG}{\ensuremath{\mathbb{G}}}
\newcommand{\pA}{\ensuremath{\mathbb{A}}}
\newcommand{\pS}{\ensuremath{\mathbb{S}}}

\newcommand{\token}[1]{\emph{#1}}

\newcommand{\sent}[1]{\emph{#1}}

\newcommand{\wsub}[2]{\scalebox{0.9}{}\ensuremath{\text{\token{#1}}_{#2}}}

\newcommand{\vect}[1]{\ensuremath{\mathbf{#1}}}

\newcommand{\trans}[1]{\textsc{#1}}

\newcommand{\transition}[2]{
	
\aboverulesep=0ex
\belowrulesep=0ex
\begin{center}
\scalebox{#1}{
\begin{tabular}{|l|l|l|l|l|}
    #2
\end{tabular}
}
\end{center}

\aboverulesep=0.4ex
\belowrulesep=0.4ex
}

\newcommand{\BERT}{$^{\mathbf{\spadesuit}}$}
\newcommand{\lastACL}{L'19}

\newcommand{\nulltag}{\ensuremath{\bot}}
\newcommand{\myedge}[3]{\ensuremath{#1 \xrightarrow{#2} #3}}

\newcommand{\myref}[1]{\S\ref{#1}}
\newcommand{\myparagraph}[1]{\textbf{#1}}

\newcommand{\possl}{\ensuremath{\text{\textit{PossL}}}}

\newcommand{\dom}{\ensuremath{\mathcal{D}}}

\newenvironment{nalign}{
	\begin{equation}
		\begin{aligned}
		}{
		\end{aligned}
	\end{equation}
	\ignorespacesafterend
}

\newcommand{\bigedge}[3]{\ensuremath{#1\xrightarrow{\text{\scriptsize #3}}#2}}


%% file: introduction.tex
\section{Introduction} \label{sec:introduction}

Over the past few years, the accuracy of neural semantic parsers which
parse English sentences into graph-based semantic representations has
increased substantially
\cite{Dozat18SemanticDependency,zhang-etal-2019-broad,bertbaseline,cai20:_amr_parsin}. Most
of these parsers use a neural model which can freely predict node
labels and edges, and most of them are tailored to a specific type of
graphbank.

Among the high-accuracy semantic parsers, the \emph{AM dependency
  parser} of
\newcite{groschwitz18:_amr_depen_parsin_typed_seman_algeb} stands out
in that it implements the Principle of Compositionality from
theoretical semantics in a neural framework. By parsing into AM
dependency trees, which represent the compositional structure of the
sentence and evaluate deterministically into graphs, this parser can
abstract away surface details of the individual graphbanks. It was the
first semantic parser which worked well across multiple graphbanks,
and set new states of the art on several of them
\cite{lindemann-etal-2019-compositional}.

However, the commitment to linguistic principles comes at a cost: the
AM dependency parser is slow. A key part of the parser is
that AM dependency trees must be \emph{well-typed} according to a type
system which ensures that the semantic valency of each word is
respected.
Existing algorithms compute all items along a parsing schema
that encodes the type constraints; they
parse e.g.\ the AMRBank at less than three tokens per
second.

In this paper, we present two fast and accurate parsing algorithms for
AM dependency trees. We first present an A* parser which
searches through the parsing schema of Groschwitz et al.'s
``projective parser'' efficiently (\myref{sec:astar}). We extend the
supertag-factored heuristic of Lewis and Steedman's
\shortcite{LewisSteedman14} A* parser for CCG with a heuristic for
dependency edge scores. This parser achieves a speed of up to 2200
tokens/s on semantic dependency parsing
\cite{OepenKMZCFHU15}, at no loss in accuracy. On AMR corpora
\cite{amBanarescuBCGGHKKPS13}, it achieves a speedup of 10x over
previous work, but still does not exceed 30 tokens/second.

We therefore develop an entirely new transition-based parser for AM
dependency trees, inspired by the stack-pointer parser of
\newcite{ma-etal-2018-stack} for syntactic dependency parsing
(\myref{sec:topdown}).
The key challenge here is to adhere to complex symbolic constraints --
the AM algebra's type system -- without running into dead ends. This
is hard for a greedy transition system and in other settings requires
expensive workarounds, such as backtracking. We ensure that our parser
avoids dead ends altogether.  We define two variants of the
transition-based parser, which choose types for words either before
predicting the outgoing edges or after, and introduce a neural model
for predicting transitions. In this way, we guarantee well-typedness
with $O(n^2)$ parsing complexity, achieve a speed of several thousand
tokens per second across all graphbanks,
and even improve the parsing
accuracy over previous AM dependency parsers by up to 1.6 points F-score.





%% file: relwork.tex
\section{Related work}
\label{sec:related-work}

In
\emph{transition-based parsing}, a dependency tree is built step by
step using nondeterministic transitions. A classifier is trained to
choose transitions deterministically
\cite{Nivre08,kiperwasser16:_simpl_accur_depen_parsin_using}.
Transition-based parsing has also been used for constituency parsing
\citep{DyerKBS16} and graph parsing \cite{E17-1051}. We build most
directly upon the top-down parser of \newcite{ma-etal-2018-stack}. 
Unlike most other transition-based parsers, our parser implements hard
symbolic constraints in order to enforce well-typedness. Such
constraints can lead transition systems into dead ends,
requiring the parser to backtrack
\citep{ytrestol2011optimistic} or return partial analyses
\citep{zhang-clark-11}. Our transition system carefully avoids dead ends. 
\citet{shi-lee-2018-valency} take hard valency constraints into account in chart-based syntactic dependency parsing, avoiding dead ends by relaxing the constraints slightly in practice.

A* parsing is a method for speeding up agenda-based chart parsers,
which takes items off the agenda based on a heuristic estimate of
completion cost. A* parsing has been used successfully for PCFGs
\cite{KleinM03}, TAG \cite{bladier2019partial}, and other grammar
formalisms. Our work is based most closely on the CCG A* parser of
\newcite{LewisSteedman14}.

Most approaches that produce semantic graphs (see
\newcite{koller19:_graph_based_meanin_repres} for an overview) model distributions over graphs directly \citep{Dozat18SemanticDependency, zhang-etal-2019-broad, bertbaseline, cai20:_amr_parsin}, while others make use of derivation trees that compositionally evaluate to graphs \citep{groschwitz18:_amr_depen_parsin_typed_seman_algeb, chen-etal-2018-accurate,fancellu-etal-2019-semantic,lindemann-etal-2019-compositional}. AM dependency parsing belongs to the latter category.





%% file: background.tex
\section{Background} \label{sec:background}

We begin by sketching the AM dependency parser of \newcite{groschwitz18:_amr_depen_parsin_typed_seman_algeb}.

\subsection{AM dependency trees}
\label{sec:amalgebra}

\citet{groschwitz18:_amr_depen_parsin_typed_seman_algeb} use \emph{AM
  dependency trees} to represent the compositional structure of a
semantic graph. Each token is assigned a \emph{graph constant}
representing its lexical meaning; dependency labels correspond to
operations of the \emph{Apply-Modify
  (AM) algebra} \citep{graph-algebra-17,GroschwitzDiss}, which
combine graphs into bigger ones.

\begin{figure}
\includegraphics[width=\columnwidth]{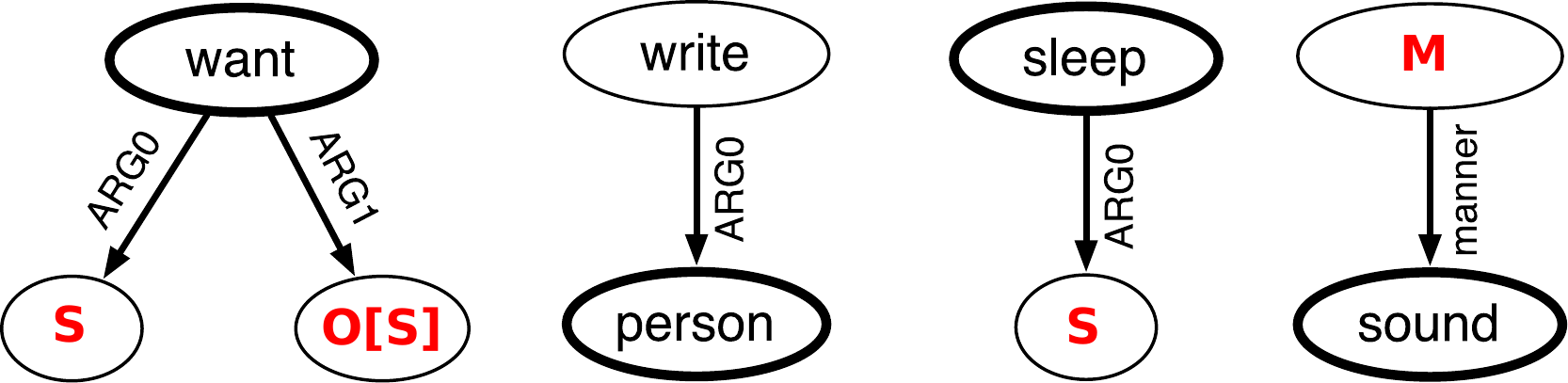}
\caption{Elementary as-graphs \G{want}, \G{writer}, \G{sleep}, and
  \G{sound}.
  }
\label{fig:as-graphs}
\end{figure}

\begin{figure*}
\includegraphics[width=\textwidth]{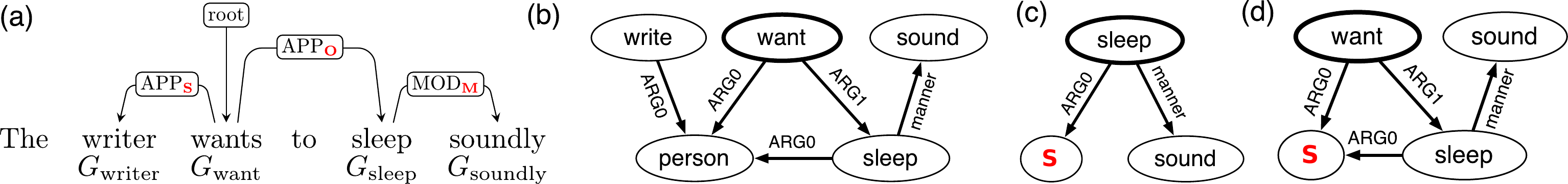}
\caption{(a) An AM dependency tree with its evaluation result (b),
  along with two partial results (c) and (d).
\label{fig:am-term}}
\end{figure*}

Fig.~\ref{fig:am-term} illustrates how an AM dependency tree (a)
evaluates to a graph (b), based on the graph constants in
Fig.~\ref{fig:as-graphs}. Each graph constant is an \emph{as-graph},
which means it has special node markers called \emph{sources}, drawn
in red, as well as a \emph{root} marked in bold. These markers are
used to combine graphs with the algebra's operations. For instance,
the \modify{\src{M}} operation in Fig.~\ref{fig:am-term}a combines the head $\G{sleep}$ with its modifier
$\G{soundly}$ by plugging the root of \G{sleep} into the \src{M}-source of
$\G{soundly}$, see (c). That is, \G{soundly} has now modified \G{sleep} and (c) is our graph for \sent{sleep soundly}. The other operation of the AM algebra, \app{}, models argument application. For example, the \app{\src{O}} operation in Fig.~\ref{fig:am-term}a plugs the root of (c) into the \src{O} source of
$\G{want}$. Note that because $\G{want}$ and (c) both have an
\src{S}-source, \app{\src{O}} merges these nodes, see (d).
The \app{\src{S}} operation then fills this \src{S}-source  with $\G{writer}$, attaching the graph with its root, to obtain the final graph in (b).\footnote{
When evaluating an AM dependency tree, the AM algebra restricts operation orders to ensure that every AM dependency tree evaluates to a unique graph.
For instance, in
Fig.~\ref{fig:am-term}, the \app{\src{O}} edge out of ``wants'' is always
tacitly evaluated before the \app{\src{S}} edge. For details on this,
we refer to
\citet{groschwitz18:_amr_depen_parsin_typed_seman_algeb} and
\citet{GroschwitzDiss}.}

\textbf{Types.} The $[\src{S}]$ annotation at the \src{O}-source of
\G{want} is a \emph{request} as to what the \emph{type} of the
\src{O}~argument of \G{want} should be. The type of an as-graph is the
set of its sources with their request annotations, so the request
$[\src{S}]$ means that the source set of the argument must be
$\{\src{S}\}$. Because this is true for (c), the AM dependency tree is
\emph{well-typed}; otherwise the tree could not be evaluated to a
graph. Thus, the graph constants lexically specify the semantic
valency of each word as well as reentrancies due to e.g.\ control, like in this example.

If an as-graph has no sources, we say it has the \emph{empty type}
$\emptytype$; if a source in a graph printed here has no annotation,
it is assumed to have the empty request (i.e.~its argument must have
no sources). We write $\tp{G}$ for the type of an as-graph $G$, and
$\req{\tau}{\alpha}$ for the request at source $\alpha$ of type
$\tau$. For example, $\req{$\tp{\G{want}}$}{\src{O}}=[\src{S}]$ and $\req{$\tp{\G{want}}$}{\src{S}}=\emptytype$. If an AM dependency (sub-)tree evaluates to
a graph $G$, we call $\tp{G}$ its \emph{term type}. For example, the sub-tree in Fig.~\ref{fig:am-term}a rooted in \sent{sleep} has term type $[\src{S}]$, since it evaluates to (c).

Below, we will build AM dependency trees by adding the outgoing edges
of a node one by one; we track types there with the following notation.
If $\tau_1$ and $\tau_2$ are the term types of AM
dependency trees $t_1,t_2$ and $\ell$ is an operation of the AM
algebra, we write $\ell\of{\tau_1,\tau_2}$ for the term type of the tree
constructed by adding $t_2$ as an $\ell$-child to $t_1$, i.e.~by adding an $\ell$-edge from the root of $t_1$ to the
root of $t_2$ (if that tree is well-typed). Intuitively, one can see this as combining a graph of type $\tau_1$ (the head) with an argument or modifier of type $\tau_2$ using operation $\ell$; the result then has type $\ell\of{\tau_1,\tau_2}$.

\subsection{AM dependency parsing} \label{sec:am-depend-pars}
\citet{groschwitz18:_amr_depen_parsin_typed_seman_algeb} approach graph parsing as first predicting a well-typed AM dependency tree for a sentence $w_1,\dots,w_n$ and then evaluating it deterministically to obtain the graph.

They train a \emph{supertag and edge-factored model}, which predicts a
\emph{supertag cost} $\graphCost{i}{G}$ for assigning a graph
constant $G$ to the token $w_i$, as well as an \emph{edge cost}
$\edgeCost{i}{j}{\oper}$ for each potential edge from word $w_i$ to
$w_j$ with label $\oper$. Tokens which are not part of the AM
dependency tree, like \sent{the} and \sent{to} in
Fig.~\ref{fig:am-term}a, are treated as if they were assigned the
special graph constant $\bot$ and an incoming \el{\lignore} edge
\edge{0}{i}{\lignore}, where $0$ represents an artificial root. The
root of the AM dependency tree (\sent{wants} in the example) is
modeled as having an incoming edge \edge{0}{i}{\lroot}.

An algorithm for AM dependency parsing searches for the well-typed AM
dependency tree which minimizes the sum of supertag and edge costs.
Finding the lowest-cost well-typed AM dependency tree for a given
sentence is NP-complete. Groschwitz et al.\ define two approximate
parsing algorithms, the \sortof{fixed tree decoder}
that fixes an unlabeled dependency tree first
, and the \sortof{projective decoder}. Our A* parser is based on the projective decoder and we focus on it here.


\textbf{Projective decoder.}
The projective decoder circumvents the NP-completeness
by searching for the best \emph{projective} well-typed AM dependency
tree. It derives parsing items using the schema \citep{shieber95:_princ_implem_deduc_parsin} shown in
Fig.~\ref{fig:rules-projective}\footnote{Originally only the fixed tree decoder used \lignore\ and \lroot\ edge scores; we extend the projective decoder here for consistency.}.

Each item encodes properties of a partial AM
dependency tree and has the form $\pitem{i}{k}{r}{\tau}{s}$, where
$[i,k] = \{j \mid i \leq j < k\}$ is the span of word indices covered
by the item, $r$ is the index of the head word, $\tau$ the type and
$s$ the cost. The Init rule assigns a supertag $G$ to a word $w_i$. The Skip-R and Skip-L
rules extend a span without changing the dependency derivation,
effectively skipping a word by assigning it the $\bot$ supertag and
drawing the corresponding \el{\lignore} edge. Finally the Arc-R and
Arc-L rules, for an AM operation $\oper$, combine two items covering
adjacent spans by drawing an edge with label $\oper$ between their
heads. Once the full chart is computed, i.e.~all items are explored, a
Viterbi algorithm yields the highest scoring well-typed AM dependency
tree.

The projective decoder has an asymptotic runtime of $O\of{n^5}$ in the
sentence length $n$.


\input rules-projective

\textbf{Notation and terminology.} Below, we assume that we obtained
three fixed non-empty finite sets in training: a set \typeSet~of types; a set \Cons~of graph
constants (the \emph{graph lexicon}) such that \typeSet~is the set of types of
the graphs in \Cons; and a set \edgeLabels~of
operations, including \lroot~and \lignore. We write \Sources~for the
set of sources occurring in \Cons~and
assume that for every source $s \in S$, $\app{s} \in \edgeLabels$.
We write $Dom(f)$ for the domain of a partial function $f$, i.e.~the set of objects for which $f$ is defined.







%% file: rules-projective.tex
\begin{figure}
  \centering 
\definecolor{LGray}{gray}{0.9}

    \small
\begin{tabular}{@{}c@{}}
\rowcolor{LGray}
  \infer{Init}{
    s = \graphCost{i}{G}
    \quad
    G \neq \bot
  }{
    \pitem{i}{i+1}{i}{\tp{G}}{s}
  }\\
  \infer{Skip-R}{\hspace{-3pt}\pitem{i}{k}{r}{\tau}{s}
    \quad
    s' = \graphCost{k}{\bot}+\edgeCost{0}{k}{\lignore}
  }{
    \pitem{i}{k+1}{r}{\tau}{s+s'}
  }\hspace{-5pt}\\
\rowcolor{LGray}
  \begin{minipage}{0.9\linewidth}
  \begin{prooftree}
  \AxiomC{\stackanchor{$\pitem{i}{k}{r}{\tau}{s}$}
                {$s' = \graphCost{i-1}{\bot}+\edgeCost{0}{i-1}{\lignore}$}
        }
  \RightLabel{Skip-L}
  \UnaryInfC{$\pitem{i-1}{k}{r}{\tau}{s+s'}$}
  \end{prooftree}
  \end{minipage}\\
  \begin{minipage}{0.9\linewidth}
  \begin{prooftree}
  \AxiomC{\stackanchor{$\pitem{i}{j}{r_1}{\tau_1}{s_1} \quad \pitem{j}{k}{r_2}{\tau_2}{s_2}$}
                {$\tau = \oper(\tau_1, \tau_2) \;\mbox{defined} \quad s = \edgeCost{r_1}{r_2}{\oper}$}
        }
  
  \RightLabel{Arc-R [\oper]}
  \UnaryInfC{$\pitem{i}{k}{r_1}{\tau}{s_1+s_2+s}$}
  \end{prooftree}
  \end{minipage}\\
\rowcolor{LGray}
  \begin{minipage}{0.9\linewidth}
      \begin{prooftree}
      \AxiomC{\stackanchor{$\pitem{i}{j}{r_1}{\tau_1}{s_1} \quad \pitem{j}{k}{r_2}{\tau_2}{s_2}$}
                    {$\tau = \oper(\tau_2, \tau_1) \;\mbox{defined} \quad s = \edgeCost{r_2}{r_1}{\oper}$}
            }
      
      \RightLabel{Arc-L [\oper]}
      \UnaryInfC{$\pitem{i}{k}{r_2}{\tau}{s_1+s_2+s}$}
      \end{prooftree}
    \end{minipage}\\
    \infer{Root}{
    \pitem{1}{n+1}{r}{\emptytype}
    \quad
    s' = \edgeCost{0}{r}{\lroot}
  }{
    \pitem{0}{n+1}{r}{\emptytype}{s+s'}
  }

  
  \end{tabular}
  \caption{Rules for the projective and A* decoder.}
  \label{fig:rules-projective}
\end{figure}


%% file: astar.tex
\section{A* AM dependency parsing} \label{sec:astar}

While the AM dependency parser yields strong accuracies across
multiple graphbanks, Groschwitz et al.'s algorithms are quite slow in
practice. For instance, the projective parser needs several hours to
parse each test set in \myref{sec:evaluation}, which seriously limits
its practical applicability. In this section, we will speed the
projective parser up through A* search.

\subsection{Agenda-based A* parsing}
\label{sec:astar-basic}

Our A* parser maintains an \emph{agenda} of parse items of the
projective parser. The agenda is initialized with the items produced by the
Init rule. Then we iterate over the agenda. In each step, we take the
item $I$ from the front of the agenda and apply the rules Skip-L and
Skip-R to it. We also attempt to combine $I$ with all previously
discovered items, organized in a parse chart, using the Arc-L and
Arc-R rules. All items thus generated are added to the agenda and the
chart. Parsing ends once we either take a goal item
$([0,n+1], r, \emptytype)$ from the agenda, or (unsucessfully) when the
agenda becomes empty.

A* parsers derive their efficiency from their ability to order the
items on the agenda effectively. They sort the agenda in ascending
order of estimated cost $f(I) = c(I) + h(I)$, where $c$ is the cost
derived for the item $I$ by the parsing rules in
Fig.~\ref{fig:rules-projective} and $h(I)$ is an \emph{outside
  estimate}. The quantity $h(I)$ estimates the difference in cost
between $I$ and the lowest-cost well-typed AM dependency tree $t$
that contains $I$. An outside heuristic is
\emph{admissible} if it is optimistic with respect to cost, i.e.\
$f(I) \leq c(t)$; in this case the parser is provably optimal, i.e.\
the first goal item which is dequeued from the agenda describes the
lowest-cost parse tree. Tighter outside estimates lead to fewer items being taken off the agenda and thus to faster runtimes.

A first \emph{trivial}, but admissible baseline lets $h(I) = 0$ for
all items $I$. This ignores the outside part and orders items purely
on their past cost. We could obtain a better outside heuristic by
following \newcite{LewisSteedman14} and summing up the cost of the
lowest-cost supertag for each token outside of the item, i.e.\
\vspace{-7pt}
$$h([i,k], r, \tau) = \sum_{j \not\in [i,k]} \min_G c(G, j).$$
\vspace{-11pt}

\noindent
This heuristic is admissible because each token will have \emph{some}
supertag selected (perhaps $\bot$) in a complete AM dependency tree,
and its cost will be equal or higher than that of the best supertag.

\subsection{Edge-based A* heuristics}
\label{sec:astar-edges}

Both of these outside heuristics ignore the fact that the cost of a
tree consists not only of the cost for the supertags,
 but also of the cost for the edges.


 We can obtain tighter heuristics by taking the edges into
 account. Observe first that the parse item determines the supertags
 and edges within its substring, and has designated one of the tokens
 as the root of the subtree it represents.
For all tokens outside of the
span of the item, the best parse tree will assign both a supertag to
the token (potentially $\nulltag$) and an incoming edge (potentially
with edge label \lroot~or \lignore). Thus, we obtain an admissible
\emph{edge-based heuristic} by adding the lowest-cost incoming edge for each outside token as follows:
$$h([i,k], r, \tau) = \sum_{j \not\in [i,k]}
  \min_G c(G, j)
  +
  \min_{\myedge{o}{\ell}{j}} c(\myedge{o}{\ell}{j})
$$

Observe finally that the edge-based heuristic is still overly
optimistic, in that it assumes that arbitrarily many nodes in the tree
may have incoming \lroot~edges (when it needs to be exactly one), and
that the choice of \lignore~and $\nulltag$ are independent (when a
node should have an incoming \lignore~edge if and only if its supertag
is $\nulltag$). We can optimize it into the \emph{ignore-aware outside
  heuristic} by restricting the $\min$ operations so they respect these
constraints.


%% file: topdown.tex
\section{Transition-based parsing}
\label{sec:topdown}

As we will see in §\ref{sec:evaluation}, the A* parser is very
efficient on the DM, PAS, and PSD corpora but still slow on EDS and AMR.

Therefore, we develop a novel transition-based algorithm for AM
dependency parsing. Inspired by the syntactic dependency parser of
\newcite{ma-etal-2018-stack}, it builds the dependency tree top-down,
starting at the root and recursively adding outgoing edges to
nodes. However, for AM dependency parsing we face an additional
challenge: we must assign a type to each node and ensure that the
overall AM dependency tree is well-typed.

We will first introduce some notation (\myref{sec:apply-sets}), then introduce three
versions of our parsing schema (\myref{sec:simplified-ltf}-\myref{sec:ltl}), give theoretical guarantees (\myref{sec:formal}) and define the neural model
(\myref{sec:training-alternativ}). 

\subsection{Apply sets}
\label{sec:apply-sets}
The transition-based parser chooses a graph constant $G_i$ for each
token $w_i$;
we call its type, $\tp{G_i}$, the \emph{lexical type} $\lextyp{l}$ of $w_i$.
As we add outgoing edges to $i$, each
outgoing $\app{\alpha}$ operation consumes the $\alpha$ source of the
lexical type. To produce a well-typed AM dependency tree of term type
$\ttyp{t}$, the sources of outgoing $\app{}$ edges at $i$ must
correspond to exactly the \emph{apply set}
$\applyset\of{\lextyp{l}, \ttyp{t}}$, which is defined as the set
$O = \{o_1,\ldots,o_n\}$ of sources such that
	$$\app{o_n}(\dots\app{o_2}(\app{o_1}(\lextyp{l},\ttyp{t}_1),\ttyp{t}_2),\dots,\ttyp{t}_n) = \ttyp{t}$$
for some types $\ttyp{t}_1,\ldots, \ttyp{t}_n$. That is, the apply set $\applyset\of{\lextyp{l}, \ttyp{t}}$ is the set of sources we need to consume to turn \lextyp{l} into \ttyp{t}.

Note that there are pairs of types for which no such set of sources exists;
e.g.~the apply set $\applyset(\emptytype,[s])$ is not defined.
In that case, we say that $[s]$ is not
\emph{apply-reachable} from $\emptytype$; the term type must always be apply-reachable from the lexical type in a well-typed tree.

\input{simple-ltf}

\input{ltf}

\input{ltl}
\input{guarantees}





\input{training-combined}



%% file: simple-ltf.tex
\subsection{Lexical type first (with dead ends)}
\label{sec:simplified-ltf}

\input{example}

We are now ready to define a first version of the transition system
for our parser.  The parser builds a dependency tree top-down and manipulates \emph{parser configurations} to track parsing decisions and ensure well-typedness. 


A parser configuration $\langle \pE, \pT, \pA, \pG, \pS \rangle$
consists of four partial functions $\pE, \pT, \pA, \pG$ that map each
token $i$ to the following:

{\hangindent=2\parindent
$\pE\of{i}$: the labeled incoming edge
of $i$, written
$\ledge{j}{\oper}{i}$, where $j$ is the head and $l$ the label;

\hangindent=2\parindent
$\pT\of{i}$: the set of possible term types at
$i$;

\hangindent=2\parindent
$\pA\of{i}$: the sources of outgoing $\app{}$ edges at
$i$, i.e.~which sources of the apply set we have covered;

\hangindent=2\parindent
$\pG\of{i}$: the graph constant at $i$.
}




These functions are partial, i.e.~they may be undefined for some
nodes.
$\pS$ is a stack of nodes that potentially still need children; we
call the node on top of $\pS$ the \emph{active node}.

The initial configuration is
$\langle \emptyset, \emptyset, \emptyset, \emptyset, \emptyset
\rangle$. A \emph{goal configuration} has an empty stack and for all
tokens $i$, it holds either that $i$ is ignored and thus has no
incoming edge, or that for some type \ttyp{t} and graph $G$, $\pT(i) = \{\ttyp{t}\}$,
$\pG(i)=G$ and
$\pA(i) = \applyset(\tp{G}, \ttyp{t})$, i.e.~$\pA(i)$ must be the apply set for the lexical type $\tp{G}$ and the term type \ttyp{t}. There must be at least one token that is not ignored.

The transition rules below read as follows:
everything above the line denotes preconditions on when the transition
can be applied; for example, that $\pT$ must map node $i$ to some set $\ttyp{T}$ of types.
The transition rule then updates the configuration by adding what is specified below the line.
An example run is shown in Fig.~\ref{fig:ltf}.

\myparagraph{\trans{Init}.} An $\trans{Init}(i)$ transition is always
the first transition and makes $i$ the root of the tree:
\transition{0.82} {
    {\small \pE} & {\small \pT } & {\small \pA} & {\small \pG} & {\small \pS}  \\
  $\emptyset$ & $\emptyset$ & $\emptyset$ & $\emptyset$ & $\emptyset$  \\
  \midrule \ledge{0}{root}{i}& $i\mapsto \{\emptytype\}$ & & & $i$ }
Fixing the term type as \emptytype~ensures that the overall evaluation result has no unfilled sources left.

\myparagraph{\trans{Choose}.}  If we have not yet chosen a graph
constant for the active node, we assign one with the
$\trans{Choose}(\ttyp{t}, G)$ transition: \transition{1.0}{
  & $i \mapsto \ttyp{T}$ &  & $i \notin Dom(\pG)$ & $\sigma|i$ \\
  \midrule
      & $i \mapsto \{\ttyp{t}\}$ & $i \mapsto \emptyset$ & $i \mapsto G$ & $\sigma|i$
}
This transition may only be applied if the specific term type
$\ttyp{t} \in \ttyp{T}$ is apply-reachable from the newly selected
lexical type $\tp{G}$. The \trans{Choose} operation is the only operation allowed when the active node does not have a graph constant yet; therefore, it always determines the lexical
type of $i$ \emph{first}, before any outgoing edges are added.

\myparagraph{\trans{Apply}.}
Once the term type $\ttyp{t}$ and graph $G$ of the active node $i$
have been chosen, the $\trans{Apply}(\alpha,j)$ operation can draw an $\app{\alpha}$ edge to a node $j$ that has no incoming edge, adding $j$ to the stack:
\transition{0.69}{
  $j \not\in Dom(\pE)$ & $i \mapsto \{\ttyp{t}\}$ & $i \mapsto A$ & $i \mapsto G$ & $\sigma|i$ \\
    \midrule
    \ledge{i}{\app{\alpha}}{j} & $j \mapsto \{\req{\tp{G}}{\alpha}\}$ & $i \mapsto A \cup \{\alpha\}$ &  & $\sigma|i|j$
  }
  Here $\alpha$ must be a source in the apply set
  $\applyset\of{\tp{G}, \ttyp{t}}$ but not in $\pA(i)$, i.e.~be a
  source of $G$ that still needs to be filled. 
Fixing the term type of $j$ ensures the type restriction of the $\app{\alpha}$ operation.


\myparagraph{\trans{Modify}.}
In contrast to outgoing \app{} edges, which are determined by the apply set, we can add arbitrary
outgoing \modify{} edges to the active node $i$. This is done with the transition
$\trans{Modify}(\beta, j)$, which draws a $\modify{\beta}$ edge to a
token $j$ that has no incoming edge, also adding $j$ to the stack:
\vspace{-5pt}
\transition{0.9}{
 $j \not\in Dom(\pE)$ & $i \mapsto \{\ttyp{t}\}$ & $i \mapsto A$ & $i \mapsto G$ &  $\sigma|i$ \\
  \midrule
  \ledge{i}{\modify{\beta}}{j} & $j \mapsto \ttyp{T'}$ & & & $\sigma|i|j$ \\
}
We require that $\ttyp{T'}$ is the set of all types $\tau' \in
\Omega$ such that all sources in $\tau'$ (except $\beta$) including
their requests are already present in $\tp{G}$, and
$\req{\tau'}{\beta}=\emptytype$, reflecting constraints on the
\modify{} operation in \newcite{GroschwitzDiss}. 

\myparagraph{\trans{Pop}.}
The \trans{Pop} transition decides that an active node that has all of
its \app{} edges will not receive any further outgoing edges, and
removes it from the stack.

\transition{0.9}{
 & $i \mapsto \{\ttyp{t}\}$ & $i \mapsto \applyset(\tp{G}, \ttyp{t})$ & $i \mapsto G$ &  $\sigma|i$ \\
  \midrule
  & & & & $\sigma$
}

%% file: example.tex
\newcommand{\colt}[1]{\textcolor{blue}{#1}}
\newcommand{\collex}[1]{#1}
\newcommand{\colls}[1]{#1}
\definecolor{LGray}{gray}{0.9}

\begin{figure*}[t]
    \centering
    \resizebox{0.99\linewidth}{!}{
    \begin{tabular}{lllllll}
    \toprule
         {\small Step} & {\small \pE} & {\small \pT} & {\small \pA} & {\small \pG} & {\small \pS} & {\small Transition} \\
         \midrule
         1 &$\emptyset$ & $\emptyset$ & $\emptyset$ & $\emptyset$ & $[]$ &  \\
         \rowcolor{LGray} 2 & \ledge{0}{\lroot}{\wsub{wants}{3}} &
                                                                   $\wsub{wants}{3} \mapsto \{\colt{\emptytype}\}$ & & & 3 & \trans{Init} 3  \\
         3 & &  & $\wsub{wants}{3} \mapsto \emptyset$ & $\wsub{wants}{3} \mapsto \G{want}$ & 3 & \trans{Choose} \colt{\emptytype}, $\langle \G{want}, \collex{[s, o[s]]} \rangle$  \\
         \rowcolor{LGray} 4 & \ledge{\wsub{wants}{3}}{\app{s}}{\wsub{writer}{2}} & $\wsub{writer}{2} \mapsto \{\colt{\emptytype}\}$ & $\wsub{wants}{3} \mapsto \{\colls{s}\}$ &  & 3 2 &  \trans{Apply} \colls{s}, 2 \\
         5 & & & $\wsub{writer}{2} \mapsto \emptyset$ & $\wsub{writer}{2} \mapsto \G{writer}$ & 3 2 & \trans{Choose} \colt{\emptytype}, $\langle \G{writer}, \collex{\emptytype} \rangle$ \\
         \rowcolor{LGray} 6 & &  &  &  & 3 & \trans{Pop} \\
         7 & \ledge{\wsub{wants}{3}}{\app{o}}{\wsub{sleep}{5}} & $\wsub{sleep}{5} \mapsto \{\colt{[s]}\}$ & $\wsub{wants}{3} \mapsto \{s,\colls{o} \}$ & &  3 5 &  \trans{Apply} \colls{o}, 5  \\
         \rowcolor{LGray} 8 & & & $\wsub{sleep}{5} \mapsto \emptyset$ & $\wsub{sleep}{5} \mapsto \G{sleep}$ &3 5 & \trans{Choose} $\colt{[s]}$, $\langle \G{sleep}, \collex{[s]} \rangle$  \\
         9 & \ledge{\wsub{sleep}{5}}{\modify{m}}{\wsub{soundly}{6}} & $\wsub{soundly}{6} \mapsto \{\colt{[\colls{m}]}, \colt{[s, \colls{m}]}\}$ & & & 3 5 6 & \trans{Modify} \colls{m}, 6  \\
         \rowcolor{LGray} 10 & & $\wsub{soundly}{6}\mapsto \{\colt{[m]}\}$ & $\wsub{soundly}{6} \mapsto \emptyset$ & $\wsub{soundly}{6} \mapsto \G{soundly}$ & 3 5 6 & \trans{Choose} $\colt{[m]}$, $\langle \G{soundly}, \collex{[m]} \rangle$  \\
         11 &  &  &  & & $[]$ & 3 $\times$ \trans{Pop} \\
         \bottomrule
    \end{tabular}
    }
    \vspace{-8pt}
    \caption{Derivation with LTF of the AM dependency tree in Fig.~\ref{fig:am-term}. The steps show only what changed for
      $\pE, \pT, \pA$ and $\pG$; the stack $\pS$
      is shown in full.
      The chosen graph constants are
      annotated with their lexical types.}
    \label{fig:ltf}
    \vspace{-15pt}
\end{figure*}

%% file: ltf.tex
\subsection{Lexical type first (without dead ends)}
\label{sec:ltf}

While the above parser guarantees well-typedness when it completes, it
can still get stuck. This is because when we \trans{Choose} a term
type $\ttyp{t}$ and lexical type $\lextyp{l}$ for a node, we
\emph{must} perform \trans{Apply} transitions for all sources in their
apply set $\applyset\of{\lextyp{l}, \ttyp{t}}$ to reach a goal
configuration. But every \trans{Apply} transition adds an incoming
edge to a token that did not have one before; if our choices for
lexical and term types require more \trans{Apply} transitions
than there are tokens without incoming edge left, the parser cannot reach a
goal configuration.

To avoid this situation, we track for each configuration $c$ the
difference $W_c - O_c$ of the number $W_c$ of tokens without an
incoming edge and the number $O_c$ of \trans{Apply} transitions we
\sortof{owe} to fill all sources. $O_c$ is obtained by summing across
all tokens $i$ the number $O_c(i)$ of \app{} children $i$ still
needs. To generalize to cases in \myref{sec:ltl} where we may not yet know the graph
constant for $i$, we let $K_c(i) = \{\tp{\pG_c(i)}\}$ if
$i \in Dom(\pG_c)$ and $K_c(i) = \Omega$ otherwise. That is, if the graph constant $\pG_c(i)$ is not yet defined, we assume we can choose it freely later. Then we can define
\vspace{-5pt}
\begin{align*}
    O_c(i) = \min_{\lextyp{l} \in K_c(i), \ttyp{t} \in \pT_c(i)}
    |\applyset(\lextyp{l}, \ttyp{t}) - \pA_c(i)|,
\end{align*}
\vspace{-16pt}

\noindent
i.e.~$O_c(i)$ is the minimal number of sources we need in addition to the ones already covered in $\pA_c(i)$ in order to cover the apply set $\applyset(\lextyp{l}, \ttyp{t})$, assuming we choose the lexical type \lextyp{l} and term type \ttyp{t} optimally within the current constraints. If $\pT$ or $\pA$ is not defined for $i$, we let $O_c(i)=0$.

Finally, given a type $\ttyp{t}$, an upper bound $n$, and a set $A$ of
already-covered sources, we let $\possl(\ttyp{t}, A, n)$ be the set of
lexical types $\lextyp{l}$ such that
$A \subseteq \applyset\of{\lextyp{l},\ttyp{t}}$ and we can reach
$\ttyp{t}$ from $\lextyp{l}$ with \app{} operations for the sources in
$A$ and at most $n$ additional \app{} operations, i.e.\
$|\applyset(\lextyp{l},\ttyp{t}) - A| \leq n$.

We prevent dead ends (see \myref{sec:formal}) by requiring that
$\trans{Choose}(\ttyp{t},
G)$ can only be applied to a configuration $c$ if $\tp{G} \in
\possl(\ttyp{t}, \emptyset, W_c-O_c)$. Then
$\ttyp{t}$ is apply-reachable from $\tp{G}$ with at most
$W_c-O_c$ \trans{Apply} transitions; this is exactly as many as we can
spare.  The
$\trans{Modify}$ transition reduces the number of tokens that have no
incoming edge without performing an
$\trans{Apply}$ transition, so we only allow it when we have tokens
\sortof{to spare}, i.e.\ $W_c-O_c \geq 1$.

%% file: ltl.tex
\subsection{Lexical type last}
\label{sec:ltl}
The lexical type first transition system chooses the graph constant
for a token early, and then chooses outgoing \app{} edges that fit the
lexical type. But of course the decisions on lexical type and outgoing
edges interact. Thus we also consider a transition system in which the
lexical type is chosen \emph{after} deciding on the outgoing edges.



\myparagraph{\trans{Apply} and \trans{Modify}.}  We modify the
\trans{Apply} and \trans{Modify} operations from \myref{sec:ltf} such
that they no longer assign term types to children and do not push the
child on the stack. This allows the transition system to add outgoing
edges to the active node $i$ without committing to types. 
The
$\trans{Apply}(\alpha,j)$ transition becomes\vspace{-0pt}
\transition{0.95}{
     $j \not\in Dom(\pE)$ & $i \mapsto \ttyp{T}$ & $i \mapsto A$ & & $\sigma|i$ \\
    \midrule
    \ledge{i}{\app{\alpha}}{j}   &  & $i \mapsto A \cup \{\alpha\}$ & & $\sigma|i$ \\
}

Because we do not yet know the types for $i$ and thus neither the apply set
$\applyset\of{\lextyp{l},\ttyp{t}}$, we cannot directly check that this
\trans{Apply} transition will not lead to a dead end. Instead, we check if
there are \emph{possible} types $\ttyp{t}$ and $\lextyp{l}$ with
$\alpha$ in their apply set, by requiring that
$\bigcup_{\ttyp{t} \in \ttyp{T}} \possl(\ttyp{t}, A \cup \{\alpha\},
W_c-1)$ is non-empty
(it is $W_c-1$ to account for the edge we are about to add).
We also keep the restriction that
$\alpha\notin A$, to avoid duplicate \app{\alpha} edges.

The $\trans{Modify}(\beta, j)$ transition becomes
\transition{1.0}{
	$j \not\in Dom(\pE)$ & & & &  $\sigma|i$ \\
	\midrule
	\ledge{i}{\modify{\beta}}{j} & & & & $\sigma|i$ \\
}
Again, we only allow it when we have tokens
\sortof{to spare}, i.e.\ $W_c-O_c \geq 1$.

\myparagraph{\trans{Finish}.} We then replace \trans{Choose} and
\trans{Pop} with a single transition $\trans{Finish}(G)$, which
selects an appropriate graph constant $G$ for the active node $i$ and pops
$i$ off the stack, such that no more edges can be added.
%


\transition{0.82}{
	$\ledge{i}{\app{\alpha_k}}{j_k}$ &&&&\\
	$\ledge{i}{\modify{\beta_k}}{l_k}$ &
	$i \mapsto \ttyp{T}$ & $i \mapsto A$ & & $\sigma|i$ \\
	\midrule
	& $i \mapsto \{\ttyp{t}\}$, & & \scalebox{0.85}{$i \mapsto G$} & $\sigma|l_1|\ldots|l_r$ \\
	& $j_k \mapsto \ttyp{T_k}$, & $j_k \mapsto \emptyset$,
	& &$\;|j_1|\ldots|j_s$  \\
	& $l_k \mapsto \ttyp{T'_k}$ & $l_k \mapsto \emptyset$ &&
}
$\trans{Finish}(G)$ is allowed if $\applyset(\tp{G}, \ttyp{t}) = A$
for some $\ttyp{t} \in \ttyp{T}$, and fixes this $\ttyp{t}$ as the
term type. In addition, \trans{Finish} pushes the child nodes $j_k$ of all
$s \geq 0$ outgoing \app{} edges onto the stack and fixes their term types as $\ttyp{T_k}=\{\req{\tp{G}}{\alpha_k}\}$ (like in \trans{Apply} of \myref{sec:simplified-ltf}). Similarly, \trans{Finish} also pushes the child nodes $l_k$ of all
$r \geq 0$ outgoing \modify{} edges onto the stack 
and computes their term type sets $\ttyp{T'_k}$ as in
the \trans{Modify} rule of \myref{sec:simplified-ltf}. 
We push the children in the reverse order of when they were created, so that they are popped off the stack in the order the edges were drawn.

Finally, since \trans{Choose} no longer exists, we must set $\pA(i)=\emptyset$ during \trans{Init}. An example run is shown in Appendix \ref{sec:appendix-am}.

%% file: guarantees.tex
\subsection{Correctness}
\label{sec:formal}


We state the main correctness results here; proofs are in Appendix \ref{sec:proofs}. We assume for all types $\lextyp{l} \in \Omega$ and all sources $\alpha \in \Sources$, that the type $\req{\lextyp{l}}{\alpha}$ is also in $\Omega$, and that for every source $\beta$ with $\modify{\beta}\in\edgeLabels$, the type $[\beta]$ is in $\Omega$. This allows us to select lexical types that do not require unexpected \app{} children.

\begin{thm}[Soundness]\label{thm:soundness}
Every goal configuration derived by LTF or LTL corresponds to a well-typed AM dependency tree.
\end{thm}

\begin{thm}[Completeness]\label{thm:completeness}
For every well-typed AM dependency tree $t$, there are sequences of LTF and LTL transitions that build $t$.
\end{thm}

\begin{thm}[No dead ends]\label{thm:noDeadEnds}
Every configuration derived by LTF or LTL can be completed to a goal configuration.
\end{thm}

%% file: training-combined.tex
\subsection{Neural model}
\label{sec:training-alternativ}

We train a neural model to predict LTF and LTL transitions, by extending Ma et al.'s stack-pointer model with means to predict graph constants and term types. We phrase AM dependency parsing as finding the most likely sequence $d^{(1)}, \ldots, d^{(N)}$ of LTF or LTL transitions given an input sentence $\vect{x}$, factorized as follows:
\vspace{-10pt}
\begin{align*}
P_{\theta}(d^{(1)}, \ldots, d^{(N)} | \mathbf{x}) &= \prod_{t=1}^N P_{\theta}(d^{(t)} | d^{(<t)}, \mathbf{x})
\end{align*}
\vspace{-12pt}

\noindent
We encode the sentence with a multi-layer BiLSTM based on embeddings for word, POS tag, lemma, named entity tag and character CNN, yielding a sequence of hidden states $\vect{s}_1, \ldots \vect{s}_n$. The decoder LSTM is initialized with the last hidden state of the encoder and is updated as follows:
$$\vect{h}^{(t)} = \text{LSTM}(\vect{h}^{(t-1)}, [\vect{s}_{tos}, \vect{s}_p, \vect{s}_c]),$$
where $tos$ denotes the node on top of the stack, $p$ the parent of $tos$ and $c$ refers to the most recently generated child of $tos$.
Let further $a^{(t)}_i \propto \exp \text{Biaffine}(\vect{h}^{(t)}, \vect{s}_i)$ be an attention score and let $\vect{s}'$ be a second BiLSTM encoding trained to predict graphs and term types.

When in the start configuration, the probability of \trans{Init} selecting node $i$ as the root is
$ P(\trans{Init } i | \vect{h}^{(t)}) = a^{(t)}_i$;
otherwise it is zero.

In LTF, if after $d^{(1)},\ldots,d^{(t-1)}$ the \trans{Choose} transition is allowed (and thus required), we have the transition probabilities

\vspace{-10pt}
\begin{small}
$$P_{\theta}(\trans{Choose }(\ttyp{t}, G) | \vect{h}^{(t)}) =  P_{\theta}(\ttyp{t} | \vect{h}^{(t)}) \cdot P_{\theta}(G | \vect{h}^{(t)})$$
\end{small}
\vspace{-15pt}

\noindent where we score the graph constant $G$ and term type \ttyp{t} with softmax functions

\vspace{-15pt}
\begin{align*}
  & P_{\theta}(G | \vect{h}^{(t)}) = \text{softmax}(\text{MLP}^{G}([\vect{h}^{(t)}, \vect{s}'_{tos}]))_{G}\\
  & P_{\theta}(\ttyp{t} | \vect{h}^{(t)}) = \text{softmax}(\text{MLP}^{tt}([\vect{h}^{(t)}, \vect{s}'_{tos}]))_{\ttyp{t}}.
\end{align*}
\vspace{-20pt}

\noindent In this situation, the probabilities of all other transitions are 0.

In contrast, if in LTF the \trans{Choose} transition is not allowed, we can draw an edge or \trans{Pop}. We score the target $j$ of the outgoing edge with the attention score $a^{(t)}_j$ and model the probability for \trans{Pop} with an artificial word at position 0 (using an attention score $a^{(t)}_0$). In other words, we have

\vspace{-15pt}
\begin{align*}
    & P_{\theta}(\ell, j | \vect{h}^{(t)}) = a^{(t)}_j \cdot P_{\theta}(\ell | \vect{h}^{(t)},  tos \rightarrow j)\\
    & P_{\theta}(\trans{Pop} | \vect{h}^{(t)}) = a^{(t)}_0
\end{align*}
\vspace{-15pt}

\noindent
where we score the edge label $\ell$ with a softmax:

\vspace{-10pt}
\begin{small}
\begin{align*}
  & P_{\theta}(\ell | \vect{h}^{(t)}, tos \rightarrow j) = \text{softmax}(\text{MLP}^{lbl}([\vect{h}^{(t)}, \vect{s}_j]))_{\ell}.
\end{align*}
\end{small}
\noindent In this situation, \trans{Choose} has probability 0.

In LTL, we must decide between drawing an edge and \trans{Finish}; we score edges as in LTF and replace the probabilities for \trans{Choose} and \trans{Pop} with 
%
\begin{align*}
    & P_{\theta}(\trans{Finish}(G) | \vect{h}^{(t)}) = a^{(t)}_0 P_{\theta}(G | \vect{h}^{(t)})
\end{align*}

\noindent where $P_{\theta}(G | \vect{h}^{(t)})$ is as above.

\textbf{Training.} The training objective is MLE of $\theta$ on a corpus of AM dependency trees. There are usually multiple transition sequences that lead to the same AM dependency tree, so we follow Ma et al. and determine a canonical sequence by visiting the children in an inside-out manner.

\textbf{Inference.}
During inference, we first decide whether we have to \trans{Choose}. If not, 
we divide each transition into two greedy decisions: we first decide, based on $a_i^{(t)}$, whether to \trans{Finish}/\trans{Pop} or whether to add an edge (and where);
second we find the graph constant (in case of \trans{Finish}) or the edge label.
To ensure well-typedness, we set the probability of forbidden transitions to 0.

\textbf{Run-time complexity.}
The run-time complexity of the parser is $O(n^2)$: $O(n)$ transitions, each of which requires evaluating attention over $n$ tokens.

The code is available at \url{https://github.com/coli-saar/am-transition-parser}.





%% file: evaluation.tex
\section{Evaluation}
\label{sec:evaluation}

\textbf{Data.} We evaluate on the DM, PAS, and PSD graphbanks from the
SemEval 2015 shared task on Semantic Dependency Parsing (SDP,
\citet{OepenKMZCFHU15}), the EDS corpus \citep{OpenSDP} and the
AMRBank releases LDC2015E86, LDC2017T10 and LDC2020T02
\cite{amBanarescuBCGGHKKPS13}. We use the AM dependency tree
decompositions of these corpora from
\citet{lindemann-etal-2019-compositional} (L'19 for short) as training
data, as well as their pre- and post-processing pipeline (including the
AMR post-processing bugfix published after submission).  We use the same
hyperparameters and hardware for all experiments (see Appendices \ref{sec:hardware} and \ref{sec:hyperparam}).

\textbf{Baselines.} We compare against the fixed tree and projective decoders of \citet{groschwitz18:_amr_depen_parsin_typed_seman_algeb}, using costs computed by the model of L'19. For the projective decoder we train with the edge existence loss recommended by \citet{groschwitz18:_amr_depen_parsin_typed_seman_algeb}. The models use pretrained
BERT embeddings \citep{devlin2018bert} without finetuning.

\input accuracy

\subsection{A* parsing}

Table~\ref{tab:st} compares the parsing accuracy of the A* parser
(with the cost model of the projective parser) across the six
graphbanks (averaged over 4 training runs of the model), with the Init
rule restricted to the six lowest-cost graph constants per token. We
only report one accuracy for A* because A* search is optimal, and thus
the accuracies with different admissible heuristics are the same. As
expected, the accuracy is on par with L'19's parser; it is slightly
degraded on DM, EDS and AMR, perhaps because these graphbanks require
non-projective AM dependency trees for accurate parsing.

\input runtimes

Parsing times are shown in Table~\ref{tab:times} as tokens per
second. We limit the number of items that can be dequeued from the
agenda to one million per sentence. This makes two sentences per AMR test set unparseable; they are given dummy
single-node graphs for the accuracy evaluation. The A* parser is
significantly faster than L'19's fixed-tree decoder; even more so than
the projective decoder on which it is based, with a 10x to 1000x
speedup. Each SDP test set is parsed in under a minute.

The speed of the A* parser is very sensitive to the accuracy of the
suppertagging model: if the parser takes many supertags for a token
off the agenda before it finds the goal item for a well-typed tree, it
will typically deqeueue many items altogether. On the SDP corpora, the
supertagging accuracy on the dev set is above 90\%; here even the
trivial heuristic is fast because it simply dequeues the best supertag
for most tokens. On AMR, the supertagging accuracy drops to 78\%;
as a consequence, the A* parser is slower overall, and the more
informed heuristics yield a higher speedup. EDS is an outlier, in that
the supertagging accuracy is 94\%, but the parser still dequeues
almost three supertags per token on average. Why this is the case
bears further study.

\subsection{Transition-based parsing}
To evaluate the transition-based parser, we extract the graph lexicon
and the type set $\Omega$ from the training and development sets such
that $\Omega$ includes all lexical types and term types used. We
establish the assumptions of \myref{sec:formal} by automatically
adding up to 14 graph constants per graphbank, increasing the graph
lexicon by less than 1\%.

The LTL parser is accurate with greedy search and parses each test set
in under a minute on the CPU and within 20 seconds on the GPU\footnote{See \citet{LindemannMsc} for the GPU implementation.}. Since the BERT embeddings take considerable time to compute, parsing without BERT leads to a parsing speed of up to 10,000 tokens per second (see Appendix \ref{sec:additionalex}).
With beam search, LTL considerably outperforms
L'19 on AMR, matching the accuracy of the fast parser of
\citet{zhang-etal-2019-broad} on AMR 17 while outperforming it by up
to 3.3 points F-score on DM. On the other graphbanks, LTL is on par
with L'19.  When evaluated without BERT, LTL outperforms L'19 by more
than 1 point F-score on most graphbanks (see Appendix \ref{sec:additionalex}).

The LTF parser is less accurate than LTL.
Beam search reduces or even closes the gap, perhaps because it can select a better
graph constant from the beam after selecting edges.

Note that accuracy drops drastically for a variant of LTL which does not enforce type constraints
(``LTL, no types'') because up to 50\%
of the predicted AM dependency trees are not well-typed
and cannot be evaluated to a graph.
The neural model does not learn to reliably construct well-typed trees
by itself; the type constraints are crucial.


Overall, the accuracy of LTL is very similar to \lastACL, except for AMR where LTL is better. We investigated this difference in performance on AMR 17 and found that LTL achieves higher precision but its recall is worse for longer sentences\footnote{For both parsers we model the dependence of recall on sentence length with linear regression; the slopes of the two models are significantly different, $p < 0.05$.}.
We suspect this is because LTL is not explicitly penalized for leaving words out of the dependency tree and thus favors shorter transition sequences.

%% file: accuracy.tex
\begin{table*}[t]
	\resizebox{\linewidth}{!}{
	\begin{tabular}{llllllllllll}
	\toprule
		& \multicolumn{2}{c}{\textbf{DM}} & \multicolumn{2}{c}{\textbf{PAS}} & \multicolumn{2}{c}{ \textbf{PSD} } & \multicolumn{2}{c}{ \textbf{EDS }} & \textbf{AMR 15} & \textbf{AMR 17} & \textbf{AMR 20}  \\
		& id F & ood F & id F & ood F & id F & ood F & Smatch F & EDM & Smatch F & Smatch F & Smatch F  \\
		\midrule
		\citet{bertbaseline}\BERT & \textbf{94.6} & 90.8 & \textbf{96.1} & \textbf{94.4} & \textbf{86.8} & 79.5 & - & - & - & - & -\\
        \citet{chen-etal-2018-accurate}   & - & - & - & - & - & - & \textbf{90.9} & \textbf{90.4} & - & - & - \\ 
        \citet{cai20:_amr_parsin}\BERT & - & - & - & - & - & - &  &  & - & \textbf{80.2} & -  \\
        \citet{zhang-etal-2019-broad}\BERT & 92.2 & 87.1 & - & - & - & - & - & - & - & 77.0\cin{0.1} & - \\
		FG'20\BERT  & 94.4 & \textbf{91.0} & 95.1 & 93.4 & 82.6 & \textbf{82.0} & - & -& - & - & - \\
		\lastACL\BERT, w/o MTL  & 93.9\cin{0.1} & 90.3\cin{0.1} &94.5\cin{0.1} & 92.5\cin{0.1} & 82.0\cin{0.1} &81.5\cin{0.3} & 90.1\cin{0.1} & 84.9\cin{0.1} & 75.4\cin{0.1} & 76.3\cin{0.2} & 75.2\cin{0.1} \\

		
        
        
          
          \midrule[0.11em]
        
          A* parser\BERT & 91.6\cin{0.1} & 88.2\cin{0.2} & 94.4\cin{0.1} & 92.6\cin{0.1} & 81.6\cin{0.1} & 81.5\cin{0.2} & 87.5\cin{0.6} & 82.8\cin{0.1} & 74.5\cin{0.1} & 75.3\cin{0.1} & 74.5\cin{0.1} \\[0.9ex]
         
			
			LTL\BERT, no types & 88.5\cin{0.3} & 82.9\cin{0.4} & 88.3\cin{0.8} & 83.6\cin{0.9} & 67.2\cin{0.6} & 67.3\cin{0.7} & 80.5\cin{0.2} & 76.3\cin{0.2} & 39.5\cin{0.5} & 46.9\cin{1.0} & 45.7\cin{1.0} \\

			LTL\BERT, greedy & 93.7\cin{0.2} & 90.0\cin{0.1} & 94.6\cin{0.2} & 92.5\cin{0.2} & 81.4\cin{0.2} & 80.7\cin{0.2} & 90.2\cin{0.1} & 85.0\cin{0.0} & 74.9\cin{0.3} & 76.5\cin{0.1} & 76.0\cin{0.1} \\
			$\quad$ beam=3 & \textbf{93.9}\cin{0.1} & 90.4\cin{0.0} & \textbf{94.7}\cin{0.1} & \textbf{92.7}\cin{0.2} & \textbf{81.9}\cin{0.1} & \textbf{81.6}\cin{0.1} & \textbf{90.4}\cin{0.0} & \textbf{85.1}\cin{0.0} & \textbf{75.7}\cin{0.3} & \textbf{77.1}\cin{0.1} & \textbf{76.8}\cin{0.1} \\[0.9ex]
            
			LTF\BERT, greedy & 92.5\cin{0.1} & 88.4\cin{0.2} & 94.0\cin{0.2} & 91.5\cin{0.2} & 77.7\cin{0.4} & 76.5\cin{0.5} & 88.0\cin{0.3} & 83.0\cin{0.3} & 71.4\cin{0.2} & 73.2\cin{0.4} & 72.6\cin{0.2} \\
			$\quad$beam=3 & \textbf{93.9}\cin{0.1} & \textbf{90.5}\cin{0.1} & 94.6\cin{0.2} & 92.6\cin{0.1} & 81.3\cin{0.1} & 80.8\cin{0.1} & 90.0\cin{0.1} & 84.8\cin{0.1} & 74.8\cin{0.2} & 76.1\cin{0.2} & 75.3\cin{0.4} \\
        \bottomrule
	\end{tabular}}
      \strut\\[-6ex]
\caption{Semantic parsing accuracies (id = in domain test set; ood = out of domain test set). \BERT\  marks models using BERT. \lastACL\ are results of \citet{lindemann-etal-2019-compositional} with fixed tree decoder (incl.~post-processing bugfix). FG'20 is \citet{lys-2020-transition}.}
\vspace{-10pt}
\label{tab:st}
\end{table*}


%% file: runtimes.tex
\begin{table}[t]
    \centering
    \resizebox{\linewidth}{!}{
    \begin{tabular}{lrrrrrrr}
        \toprule
        & \textbf{DM} & \textbf{PAS} & \textbf{PSD} & \textbf{EDS} & \textbf{A15} & \textbf{A17} & \textbf{A20} \\
        \midrule
        projective, \lastACL\BERT costs
        & 3 & 2 & 4 & 4 & $<$2 & $<$2 & $<$2 \\
        \lastACL\BERT fixed tree & 710   & 97   & 265 & 542 & $<$4 & $<$3 & $<$3 \\
        \midrule
               
        A*\BERT, trivial & 706 & 2096 & 1235 & 105 & $<$9 & $<$10 & $<$6 \\
        A*\BERT, edge-based & 725 & 2105 & 1421 & 129 & $<$20 & $<$26 & $<$20 \\
        A*\BERT, ignore-aware & 712 & 2167 & 1318 & 136 & $<$22 & $<$30 & $<$26 \\
        \midrule
        LTL\BERT, GPU, greedy & \textbf{4,750} & \textbf{4,570} & \textbf{2,742} & \textbf{4,443} & \textbf{1,977} & \textbf{2,116} & \textbf{1,946} \\
        LTL\BERT, CPU, greedy & 1,094& 913 & 1,126 & 968& 879 & 962 & 865 \\
$\quad$ beam=3 & 241& 203 & 231 & 217 & 217 & 205 & 198 \\
LTF\BERT, CPU, greedy & 852 & 791 & 688 & 673 & 563 & 424 & 514 \\
$\quad$ beam=3 & 145 & 123 & 96 & 108 & 100 & 76 & 78\\
        \bottomrule
        
    \end{tabular}}
      \strut\\[-4ex]
    \caption{Avg.\  parsing speed in tokens/s on test sets. $<$ indicates where parsing was interrupted due to timeout. }
    
    \label{tab:times}
\end{table}

%% file: conclusion.tex
\section{Conclusion} \label{sec:conclusion}

We have presented two fast and accurate algorithms for AM dependency
parsing: an A* parser which optimizes Groschwitz et al.'s projective
parser, and a novel transition-based parser which builds an AM
dependency top-down while avoiding dead ends. 

The parsing speed of the A* parser differs dramatically for the different graphbanks. In contrast, the parsing speed with the transition systems is less sensitive to the graphbank and faster overall. The transition systems also achieve higher accuracy.

In future work, one could make the A* parser more accurate by
extending it to non-projective dependency trees, especially on DM,
EDS and AMR. The transition-based parser could be made more accurate by making
bottom-up information available to its top-down choices, e.g.\ with
Cai and Lam's \shortcite{cai20:_amr_parsin} ``iterative inference''
method. 
It would also be interesting to see if our method for avoiding dead ends can be applied to other formalisms with complex symbolic restrictions.



%% file: appendix-results.tex
\section{Additional experiments and dev accuracies}
\label{sec:additionalex}
Table \ref{tab:fullresults} shows the results of further experiments (means and standard deviations over 4 runs). Models that do not use BERT, use GloVe embeddings of size 200. Note that we use the pre- and postprocessing of \citet{lindemann-etal-2019-compositional} in the most recent version, which fixes a bug in AMR post-processing\footnote{see {\small \href{https://github.com/coli-saar/am-parser}{https://github.com/coli-saar/am-parser}}}.

For each model trained, Table \ref{tab:devresults} shows the performance of one run on the development set.

\begin{table*}[t]
	\resizebox{\linewidth}{!}{
		\begin{tabular}{llllllllllll}
			\toprule
			& \multicolumn{2}{c}{\textbf{DM}} & \multicolumn{2}{c}{\textbf{PAS}} & \multicolumn{2}{c}{ \textbf{PSD} } & \multicolumn{2}{c}{ \textbf{EDS }} & \textbf{AMR 15} & \textbf{AMR 17} & \textbf{AMR 20} \\
			& id F & ood F & id F & ood F & id F & ood F & Smatch F & EDM & Smatch F & Smatch F & Smatch F  \\
			\midrule
			
			
			\lastACL\ + CharCNN & 90.5 \cin{0.1} & 84.5 \cin{0.1} & 91.5\cin{0.1}&		86.5\cin{0.1}	 &	78.4\cin{0.2}	&	74.8\cin{0.2}& 87.7\cin{0.1} & 82.8\cin{0.1} & 70.5\cin{0.4} & 71.7\cin{0.2} & 70.4\cin{0.5} \\
			
			\lastACL \BERT + CharCNN & 93.8\cin{0.1}&90.2\cin{0.1}&94.6\cin{0.1} &		92.5\cin{0.1}	&	\textbf{81.9}\cin{0.1}	&	81.5\cin{0.2}&90.2\cin{0.1}& 85.0\cin{0.1}&75.4\cin{0.2}&76.4\cin{0.1}  & 74.8\cin{0.2} \\

			\midrule
			
            A* parser\BERT & 91.6\cin{0.1} & 88.2\cin{0.2} & 94.4\cin{0.1} & 92.6\cin{0.1} & 81.6\cin{0.1} & 81.5\cin{0.2} & 87.5\cin{0.6} & 82.8\cin{0.1} & 74.5\cin{0.1} & 75.3\cin{0.1} & 74.5\cin{0.1}\\
			\midrule

            LTL\BERT, no types & 88.5\cin{0.3} & 82.9\cin{0.4} & 88.3\cin{0.8} & 83.6\cin{0.9} & 67.2\cin{0.6} & 67.3\cin{0.7} & 80.5\cin{0.2} & 76.3\cin{0.2} & 39.5\cin{0.5} & 46.9\cin{1.0} & 45.7\cin{1.0} \\

			LTL, greedy & 91.4\cin{0.2} & 85.4\cin{0.1} & 92.5\cin{0.1} & 87.9\cin{0.1} & 78.6\cin{0.2} & 74.9\cin{0.3} & 88.2\cin{0.1} & 83.0\cin{0.1} & 71.1\cin{0.2} & 73.1\cin{0.2}  & 72.5\cin{0.2} \\
			$\quad$ beam = 3 & 91.5\cin{0.2} & 86.0\cin{0.1} & 92.7\cin{0.2} & 88.3\cin{0.3} & 79.4\cin{0.2} & 76.2\cin{0.2} & 88.3\cin{0.2} & 83.1\cin{0.1} & 71.8\cin{0.3} & 73.7\cin{0.3} & 73.3\cin{0.3} \\

            LTL\BERT, greedy & 93.7\cin{0.2} & 90.0\cin{0.1} & 94.6\cin{0.2} & 92.5\cin{0.2} & 81.4\cin{0.2} & 80.7\cin{0.2} & 90.2\cin{0.1} & 85.0\cin{0.0} & 74.9\cin{0.3} & 76.5\cin{0.1} & 76.0\cin{0.1} \\
			$\quad$ beam=3 & \textbf{93.9}\cin{0.1} & 90.4\cin{0.0} & \textbf{94.7}\cin{0.1} & \textbf{92.7}\cin{0.2} & \textbf{81.9}\cin{0.1} & \textbf{81.6}\cin{0.1} & \textbf{90.4}\cin{0.0} & \textbf{85.1}\cin{0.0} & \textbf{75.7}\cin{0.3} & \textbf{77.1}\cin{0.1} & \textbf{76.8}\cin{0.1} \\
            
            \midrule
            
            LTF\BERT, no types & 85.0\cin{0.3} & 78.0\cin{0.7} & 86.9\cin{0.6} & 81.4\cin{0.4} & 63.1\cin{1.4} & 62.5\cin{0.7} & 72.4\cin{0.4} & 69.1\cin{0.3} & 30.8\cin{0.3} & 36.6\cin{2.7} & 37.6\cin{0.7}
 \\
                        
			LTF, greedy & 89.7\cin{0.4} & 83.0\cin{0.3} & 91.8\cin{0.2} & 86.6\cin{0.2} & 74.2\cin{0.4} & 69.8\cin{0.6} & 86.1\cin{0.1} & 81.3\cin{0.1} & 67.7\cin{0.2} & 69.2\cin{0.3} & 69.0\cin{0.2} \\
			$\quad$ beam = 3 & 91.5\cin{0.3} & 85.6\cin{0.2} & 92.6\cin{0.2} & 88.0\cin{0.1} & 78.8\cin{0.4} & 75.1\cin{0.2} & 88.1\cin{0.2} & 82.9\cin{0.1} & 71.0\cin{0.2} & 72.2\cin{0.2} & 72.2\cin{0.2} \\  
			

            LTF\BERT, greedy & 92.5\cin{0.1} & 88.4\cin{0.2} & 94.0\cin{0.2} & 91.5\cin{0.2} & 77.7\cin{0.4} & 76.5\cin{0.5} & 88.0\cin{0.3} & 83.0\cin{0.3} & 71.4\cin{0.2} & 73.2\cin{0.4} & 72.6\cin{0.2} \\
			$\quad$beam=3 & \textbf{93.9}\cin{0.1} & \textbf{90.5}\cin{0.1} & 94.6\cin{0.2} & 92.6\cin{0.1} & 81.3\cin{0.1} & 80.8\cin{0.1} & 90.0\cin{0.1} & 84.8\cin{0.1} & 74.8\cin{0.2} & 76.1\cin{0.2} & 75.3\cin{0.4} \\
			\bottomrule
	\end{tabular}}
	\caption{Fulls details of accuracies of parsers we have trained (id = in domain test set; ood = out of domain test set). \BERT\  marks models using BERT. \lastACL\ is \citet{lindemann-etal-2019-compositional} with fixed tree decoder.}
	\label{tab:fullresults}
\end{table*}


\begin{table}[t]
	\centering
	\resizebox{\linewidth}{!}{
		\begin{tabular}{lcccccccc}
			\toprule
			& \textbf{DM} & \textbf{PAS} & \textbf{PSD} & \multicolumn{2}{c}{ \textbf{EDS} } & \textbf{AMR 15} & \textbf{AMR 17} & \textbf{AMR 20} \\
			& F & F & F & Smatch & EDM & Smatch & Smatch & Smatch \\
			\midrule
			L'19 + charCNN & 91.2 & 91.7 & 80.6 & 88.6 & 84.1 & 71.6 & 72.9 &  73.0 \\
			L'19\BERT + charCNN & 94.2 & 95.0 & 84.3 & 90.6 & 86.0 & 75.9 & 77.3 & 77.5 \\
			\midrule
			A*\BERT & 92.1 & 94.6 & 84.0 & 88.0 & 83.8 & 75.1 & 76.4 & 76.9 \\
			\midrule
			LTL, greedy & 92.1 & 92.9  & 80.8  & 89.1 & 84.6  & 72.7   & 74.7 & 75.6 \\
			LTL\BERT, greedy & 94.1 & 95.1  & 83.4  & 90.6  & 85.9  & 76.1 & 78.0 & 78.4 \\
			\midrule
			LTF, greedy & 91.0  & 92.4 & 75.7  & 87.1  & 82.8 & 69.2 & 70.7 & 72.1  \\
			LTF\BERT, greedy & 92.9  & 94.59  & 79.7 & 88.7  & 84.2 & 72.8 & 74.8 & 75.4 \\
			\bottomrule
	\end{tabular}}
	\caption{Results on development sets.\BERT\  marks models using BERT. \lastACL\ is \citet{lindemann-etal-2019-compositional} with fixed tree decoder.}
	\label{tab:devresults}
\end{table}

\begin{table}[t]
\resizebox{\linewidth}{!}{
\begin{tabular}{lrrrrrrr}
	\toprule
	&\textbf{DM} & \textbf{PAS} & \textbf{PSD} & \textbf{EDS} & \textbf{AMR 15} & \textbf{AMR 17} & \textbf{AMR 20} \\
	\midrule
	LTL, greedy & 1,180 & 1,128 & 1,288 & 1,154 & 1,121 & 1,162 & 1,148\\
	LTL, beam=3 & 257& 229 & 243 & 224 & 234 & 222 & 210 \\
	LTL, GPU, greedy & \textbf{10,266}& \textbf{10,271}& \textbf{4,201}& \textbf{9,188} & \textbf{3,647} & \textbf{3,413} & \textbf{2,912} \\
	& & & & & & \\
	LTF, greedy & 957 & 908& 755 & 752& 672 & 578 & 572 \\
	LTF, beam=3 & 153 & 126 & 97 & 113 & 104 & 91 & 81 \\
	\bottomrule
\end{tabular}}
\caption{Avg.\ parsing speed of transition systems in tokens/s on (in-domain) test sets without BERT. For result with BERT, see main paper.}
\label{tab:speed-glove}
\end{table}

Table \ref{tab:smatch-compare} shows F-scores of different versions of Smatch on the AMR tests. See also Appendix \ref{sec:eval-metrics}.

\begin{table}[t]
	\resizebox{\linewidth}{!}{
		\begin{tabular}{lll|ll}
			\toprule
			& \multicolumn{2}{c}{\textbf{AMR 2015}} & \multicolumn{2}{c}{\textbf{AMR 2017}}\\
			& new F &  \lastACL\ F & new F &  \lastACL\ F \\
			\midrule

			\lastACL\ + CharCNN &  70.5\cin{0.4} & 70.2\cin{0.4} & 71.7\cin{0.2} & 71.4\cin{0.2} \\

			\lastACL \BERT + CharCNN &  75.4\cin{0.2}& 75.1\cin{0.2} & 76.4\cin{0.1} & 76.1\cin{0.1} \\
			\lastACL \BERT, w/o MTL &  75.4\cin{0.1}& 75.1\cin{0.2} & 76.3\cin{0.2} & 76.0\cin{0.2} \\
			
			\midrule 
			A* parser\BERT & 74.5\cin{0.1} & 74.2\cin{0.1} & 75.3\cin{0.1} & 75.1\cin{0.1} \\
			\midrule

			LTL, greedy & 71.1\cin{0.2} & 70.7\cin{0.2} & 73.1\cin{0.2} & 72.8\cin{0.3} \\

			$\quad$ beam = 3 & 71.8\cin{0.3} & 71.4\cin{0.3} & 73.7\cin{0.3} & 73.4\cin{0.3}  \\
			
            LTL\BERT, greedy &  74.9\cin{0.3} & 74.5\cin{0.3} & 76.5\cin{0.1} & 76.3\cin{0.1}  \\

			$\quad$ beam=3 & 75.7\cin{0.3} & 75.3\cin{0.3} & 77.1\cin{0.1} & 76.8\cin{0.1} \\
            
            \midrule
                     
			LTF, greedy &  67.7\cin{0.2}& 67.3\cin{0.2}& 69.2\cin{0.3} & 68.9\cin{0.2} \\
            
			$\quad$ beam = 3 &  71.0\cin{0.2} & 70.6\cin{0.2} & 72.2\cin{0.2}&71.9\cin{0.2} \\

			LTF\BERT, greedy &  71.4\cin{0.2} & 71.1\cin{0.1} & 73.2\cin{0.4} & 72.9\cin{0.4}\\

			$\quad$beam=3 & 74.8\cin{0.2} & 74.5\cin{0.2} & 76.1\cin{0.2} & 75.8\cin{0.2}\\
            
			\bottomrule
	\end{tabular}}
	\caption{Results on AMR test sets with different versions of Smatch. \lastACL\ F is the version that was used by \citet{lindemann-etal-2019-compositional} and \textit{new F} is version 1.0.4.}
	\label{tab:smatch-compare}
\end{table}


%% file: appendix-hyperparams.tex
\section{Hardware and parsing experiments}
\label{sec:hardware}
All parsing experiments were performed on Nvidia Tesla V100 graphics
cards and Intel Xeon Gold 6128 CPUs running at 3.40~GHz.

We measure run-time as the sum of the GPU time and the CPU time on a single core for all approaches. When computing scores for A*, we use a batch size of 512 for all graphbanks but AMR, where we use a batch size of 128. We use a batch size of 64 for LTL and LTF for parsing on the CPU. The transition probabilites are computed on the GPU and then transferred to the main memory. In the parsing experiments with LTL where the transition system is run on the GPU as well, we use a batch size of 512, except for AMR, for which we use a batch size of 256.

The A* algorithm is implemented in Java and was run on the GraalVM 20 implementation of the JVM.

We run the projective parser and the fixed tree parser of \citet{groschwitz18:_amr_depen_parsin_typed_seman_algeb} with the 6 best supertags. When parsing with the fixed tree parser is not completed with $k$ supertags within
30 minutes, we retry with $k-1$ supertags. If $k = 0$, we use a dummy graph with a single node.

LTL and LTF are implemented in python and run on CPython version 3.8.

\section{Hyperparameters and training details}
\label{sec:hyperparam}

\subsection{Scores for A*}
We obtain the scores by training the parser of \citet{lindemann-etal-2019-compositional}. Since \citet{groschwitz18:_amr_depen_parsin_typed_seman_algeb} argue that a hinge loss such as the one that \lastACL\ use might not be well-suited for the projective parser, we replaced it by the log-likelihood loss of \citet{groschwitz18:_amr_depen_parsin_typed_seman_algeb}. The development metric based on which the model is chosen is the arithmetic mean between supertagging accuracy and labeled attachment score.

We follow \citet{lindemann-etal-2019-compositional} in the hyperparamters, with two exceptions: we use batch size of 32 instead of 64 because of memory constraints and 
we add a character CNN to the model to make it more comparable with the model of the transition systems; see below for its hyperparameters. In order to tease apart the impact of the character CNN, we include the performance of a \lastACL\  model with the character CNN in Table \ref{tab:fullresults}. Differences are within one standard deviation of the results obtained with the original architecture used in \lastACL.

\subsection{LTL and LTF}
We set the hyperparameters manually without extensive hyperparameter search, mostly following \citet{ma-etal-2018-stack}. We followed \citet{lindemann-etal-2019-compositional} for number of hidden units and dropout in the MLPs for predicting graph constants and for the size of embeddings.

We follow \citet{lindemann-etal-2019-compositional} in splitting the prediction of a graph constant into predicting a delexicalized graph constant and a lexical label.

We train all LTL and LTF models for 100 epochs with Adam using a batch size of 64. We follow \citet{ma-etal-2018-stack} in setting $\beta_1, \beta_2 = 0.9$ and the initial learning rate to $0.001$. We don't perform weight decay or gradient clipping. In experiments with GloVe, we use the vectors of dimensionality 200 (6B.200d) and fine-tune them. Following \citet{ma-etal-2018-stack}, we employ a character CNN with 50 filters of window size 3.

We use the \texttt{BERT large} version of BERT and average the layers. The weights for the average are learned but we do not fine-tune BERT itself.

For the second encoding of the input sentence, $\mathbf{s}'$, we use a single-layer bidirectional LSTM when using BERT and a two-layer bidirectional LSTM when using GloVe. On top of $\mathbf{x}'$ we perform variational dropout with $p=0.33$, as well as on top of $\mathbf{s}$ and $\mathbf{s}'$.
The other hyperparameters are listed in Tables \ref{tab:embeddings}, \ref{tab:hma} and \ref{tab:hconstants}. The number of parameters of the LTL and LTF models are in table \ref{tab:parameters}.

Training an LTL or LTF model with BERT took at most 24 hours, and about 10 hours for AMR 15. Training with GloVe is usually a two or three hours shorter.

\begin{table}[t]
    \centering
\begin{tabular}{ll}
\toprule
    POS & 32 \\
    Characters & 100 \\
    NE embedding & 16 \\
    \bottomrule
\end{tabular}
    \caption{Dimensionality of embeddings used in all experiments.}
    \label{tab:embeddings}
\end{table}

\begin{table}[t]
    \centering
\begin{tabular}{ll}
\toprule
\textbf{All LSTMs}: & \\
LSTM hidden size (per direction) & 512 \\
LSTM layer dropout & 0.33 \\
LSTM recurrent dropout & 0.33 \\
\hline
Encoder LSTM layers used for $\mathbf{s}$ & 3\\
\hline
Decoder LSTM layers & 1 \\
\hline 
\textbf{MLPs before bilinear attention} & \\
Layers & 1 \\
Hidden units & 512 \\
Activation & elu \\
Dropout & 0.33 \\
\hline
\textbf{Edge label model} & \\
Layers & 1 \\
Hidden units & 256 \\
Activation & tanh \\
Dropout & 0.33 \\
\bottomrule
\end{tabular}
    \caption{Hyperparameters of LTL and LTF}
    \label{tab:hma}
\end{table}

\begin{table}[t]
    \centering
    \begin{tabular}{ll}
    \toprule
    Layers & 1 \\
    Hidden units & 1024 \\
    Activation & tanh \\
    Dropout & 0.4 \\
    \bottomrule
\end{tabular}
    \caption{Hyperparameters used in MLPs for predicting delexicalized constants, term types and lexical labels.}
    \label{tab:hconstants}
\end{table}

\begin{table}
\resizebox{\linewidth}{!}{
\begin{tabular}{lrrrr|rr}
\toprule
& \multicolumn{2}{c}{LTL} & \multicolumn{2}{c}{LTF} & \multicolumn{2}{c}{\lastACL} \\
& GloVe & BERT & GloVe & BERT & Glove & BERT \\
\midrule
DM & 67.39 & 61.77 & 69.59 & 63.97 & 19.19 & 8.76 \\
PAS & 66.71 & 61.05 & 68.90 & 63.24 & 18.54 & 8.05 \\
PSD & 73.95 & 68.15 & 76.40 & 70.60 & 25.84 & 15.15 \\
EDS & 70.35 & 65.98 & 72.59 & 68.23 & 21.52 & 12.97 \\
AMR 15 & 71.49 & 68.34 & 73.88 & 70.73 & 22.07 & 15.34 \\
AMR 17 & 76.42 & 71.60 & 78.86 & 74.04 & 27.84 & 18.61 \\
AMR 20 & 82.63 & 75.56 & 85.13 & 78.06 & 35.16 & 22.33  \\ 
\bottomrule
\end{tabular}

}
\caption{Number of trainable parameters (including GloVe embeddings) in millions.}
\label{tab:parameters}
\end{table}

\section{Data}
We use the AM dependency trees of \citet{lindemann-etal-2019-compositional} as training data, along with their pre-processing. See their supplementary materials for more details.
For completeness, Table \ref{tab:stats} shows the number of AM dependency trees in the training sets as well as the number of sentences and tokens in the test sets. Note that the heuristic approach cannot obtain AM dependency trees for all graphs in the training data but nothing is left out of the test data.

\begin{table}
	\resizebox{\linewidth}{!}{
	\begin{tabular}{lcc|cc}
		\toprule
		& \multicolumn{2}{c}{ Training } & \multicolumn{2}{c}{ Test } \\
		\midrule
		& Sentences & AM dep. trees & Sentences & Tokens \\
		DM & 35,657 & 31,349 & 1,410 & 33,358 \\
		PAS & 35,657 & 31,796 & 1,410 & 33,358 \\
		PSD & 35,657 & 32,807 & 1,410 & 33,358 \\
		EDS & 33,964 & 25,680 & 1,410 & 32,306 \\
		AMR 15 & 16,833 & 15,472 & 1,371 & 28,458 \\
		AMR 17 & 36,521 & 33,406 & 1,371 & 28,458 \\
        AMR 20 & 55,635	& 51.515 & 1,898 & 36,928 \\
		\bottomrule
	\end{tabular}}
	\caption{Data statics after preprocessing.Test set is in-domain for SDP.}
	\label{tab:stats}
\end{table}

We use the standard splits on all data sets into training/dev/test, again following \citet{lindemann-etal-2019-compositional}.

PAS, PSD and AMR are licensed by LDC but the DM and EDS data can be downloaded from \href{http://hdl.handle.net/11234/1-1956}{http://hdl.handle.net/11234/1-1956}.

\section{Evaluation metrics}
\label{sec:eval-metrics}
\paragraph{DM, PAS and PSD} 
We compute labeled F-score with the evaluation toolkit that was developed for the shared task: \href{https://github.com/semantic-dependency-parsing/toolkit}{https://github.com/semantic-dependency-parsing/toolkit}.

\paragraph{EDS}
We evaluate with Smatch \citet{CaiK13}, in this implementation due to its high speed:  {\tiny \href{https://github.com/Oneplus/tamr/tree/master/amr\_aligner/smatch}{github.com/Oneplus/tamr/tree/master/amr\_aligner/smatch}} and
EDM \citep{dridan2011parser} in the implementation of \citet{BuysBlunsom17}: \href{https://github.com/janmbuys/DeepDeepParser}{https://github.com/janmbuys/DeepDeepParser}. We follow \citet{lindemann-etal-2019-compositional} in using Smatch as development metric.

\paragraph{AMR}
We evaluate with Smatch in the original implementation \href{https://github.com/snowblink14/smatch}{https://github.com/snowblink14/smatch}. 
In the main paper, we report results with Smatch 1.0.4, which are somewhat better than with earlier versions. This also applies to the results of \citet{lindemann-etal-2019-compositional}. Table \ref{tab:smatch-compare} shows results with the Smatch version that were originally used in \citet{lindemann-etal-2019-compositional} (Commit \texttt{ad7e65} from August 2018).

%% file: appendix-am.tex
\section{Example for LTL}
\label{sec:appendix-am}

Fig.~\ref{fig:ltl} shows an example of a derivation with LTL, analogous to the one in Fig.~\ref{fig:ltf}.
\begin{figure*}[t]
	\centering
	\resizebox{\linewidth}{!}{
		\begin{tabular}{lllllll}
			\toprule
         Step & \pE &  \pT & \pA & \pG &  \pS &  Transition \\
			\midrule
			1 &$\emptyset$ & $\emptyset$ & $\emptyset$ & $\emptyset$ & $[]$ &  \\
			\rowcolor{LGray} 2 & \ledge{0}{\lroot}{\wsub{wants}{3}} &
			$\wsub{wants}{3} \mapsto \{\colt{\emptytype}\}$ &$\wsub{wants}{3} \mapsto \emptyset$  & & 3 & \trans{Init} 3  \\
			3 & \ledge{\wsub{wants}{3}}{\app{s}}{\wsub{writer}{2}} &  & $\wsub{wants}{3} \mapsto \{\colls{s}\}$ &  & 3 &  \trans{Apply} (\colls{s}, 2) \\
			\rowcolor{LGray} 4 & \ledge{\wsub{wants}{3}}{\app{o}}{\wsub{sleep}{5}} &  & $\wsub{wants}{3} \mapsto \{s,\colls{o} \}$ & &  3 &  \trans{Apply} (\colls{o}, 5)  \\
			\multirow{ 2}{*}{5} & &$\wsub{writer}{2} \mapsto \{\colt{\emptytype}\}$,   & $\wsub{writer}{2} \mapsto \emptyset$,   &  \multirow{ 2}{*}{$\wsub{wants}{3} \mapsto \G{want}$} & \multirow{ 2}{*}{5 2} &\multirow{ 2}{*}{\trans{Finish}($\langle \G{want}, \collex{[s, o[s]]} \rangle$)} \\
			& & $\wsub{sleep}{5} \mapsto \{\colt{[s]}\}$ & $\wsub{sleep}{5} \mapsto \emptyset$ & & \\
			\rowcolor{LGray} 6 & &  &  & $\wsub{writer}{2} \mapsto \G{writer}$ & 5 & \trans{Finish}($\langle\G{writer}, \collex{\emptytype} \rangle$) \\
			7 & \ledge{\wsub{sleep}{5}}{\modify{m}}{\wsub{soundly}{6}} &  & & & 5 & \trans{Modify} (\colls{m}, 6)  \\
			\rowcolor{LGray} 8 & & $\wsub{soundly}{6} \mapsto \{\colt{[\colls{m}]}, \colt{[s, \colls{m}]}\}$ &  &  $\wsub{sleep}{5} \mapsto \G{sleep}$ & 6 & \trans{Finish}($\langle\G{sleep}, \collex{[s]}\rangle$) \\
			9 & & $\wsub{soundly}{6} \mapsto \{\colt{[\colls{m}]}\}$  &  &  $\wsub{soundly}{6} \mapsto \G{soundly}$ &  & \trans{Finish}($\langle \G{soundly}, \collex{[m]}\rangle$) \\
			\bottomrule
		\end{tabular}
	}
	\caption{Derivation with LTL of the AM dependency tree in Fig.~\ref{fig:am-term}. The steps show only what changed for
		$\pE, \pT, \pA$ and $\pG$; the stack $\pS$
		is shown in full.
		The chosen graph constants are
		annotated with their lexical types.}
	\label{fig:ltl}
\end{figure*}

%% file: proofs-new.tex
\section{Proofs}
\label{sec:proofs}
The proofs given here follow exactly \citet{LindemannMsc}.

The transition systems LTF and LTL are designed in such a way that they enjoy three particularly important properties: soundness, completeness and the lack of dead ends.
In this section, we phrase those guarantees in formal terms, determine which assumptions are needed and prove the guarantees. It will turn out that significant assumptions are only needed to guarantee that there are no dead ends.

Throughout this section we assume the type system of \cite{GroschwitzDiss}, where types are formally defined as DAGs with sources as nodes, and requests being defined via the edges.



The definition of a goal condition is quite strict but it can be shown that for LTF and LTL simpler conditions are equivalent:
\begin{lemma}
	Let $c$ be a configuration derived by LTF.
	$c$ is a goal configuration if and only if $\pS_c$ is empty and $\pG_c$ is defined for some $i$.
	\label{lemma:empty-stack-ltf}
\end{lemma}
\begin{proof} 
	$\implies$ \\
	This follows trivially from the definition of a goal condition.
	\\
	$\impliedby$ \\
	We have to validate that for each token $l$, either $l$ is ignored and thus has no
	incoming edge, or that for some type \ttyp{t} and graph $G$, $\pT_c(l) = \{\ttyp{t}\}$,
	$\pG_c(l)=G$ and
	$\pA_c(l) = \applyset(\tp{G}, \ttyp{t})$. Additionally, there must be at least one token $j$ such that $\pT_c(j) = \{\ttyp{t}\}$,
	$\pG_c(j)=G$ and
	$\pA_c(j) = \applyset(\tp{G}, \ttyp{t})$. We first show that this latter condition holds for token $i$ for which $\pG_c$ is defined.
	Notice that $i$ must have been on the stack and a \trans{Choose} transition has been applied. Since it is not on the stack anymore, a $\trans{Pop}$ transition has been applied in some configuration $c'$ where $i$ was the active node. This means that $\pA_{c}(i) = \pA_{c'}(i) = \applyset(\tp{\pG_{c}(i)}, \ttyp{t})$ with $\pT_{c}(i) = \pT_{c'}(i) = \{\ttyp{t}\}$ and thus $i$ fulfills its part for $c$ being a goal configuration.
	
	We assumed that $c$ was derived by LTF, so let $s$ be an arbitrary transition sequence that derives $c$ from the initial state (there might be multiple).
	We can divide the tokens in the sentence into two groups, based on whether they have ever been on the stack over the course of $s$:
	\begin{itemize}
		\item let $j$ be an arbitrary token such that there is a state $c'$ produced by a prefix of the transition sequence $s$ where $j$ is on the stack. Here, the same argument holds as above: since $j$ is no longer on the stack, a $\trans{Pop}$ transition must have been applied which implies that $\pA_{c}(j) = \applyset(\tp{\pG_{c}(j)}, \ttyp{t})$ with $\pT_{c}(j) = \{\ttyp{t}\}$.
		\item let $j$ be an arbitrary token such that there is \textit{no} state $c'$ produced by a prefix of the transition sequence $s$ where $j$ is on the stack. Clearly, such a token $j$ does not have an incoming edge and thus also fulfills its part.
	\end{itemize}
\end{proof}

\begin{lemma}
	Let $c$ be a configuration derived by LTL.
	$c$ is a goal configuration if and only if $\pS_c$ is empty and $\pG_c$ is defined for some $i$.
	\label{lemma:empty-stack-ltl}
\end{lemma}
\begin{proof}
	The proof is analogous to the proof of Lemma \ref{lemma:empty-stack-ltf}.
\end{proof}

\subsection{Soundness}
An important property of the transition systems is that they are sound, that is, every AM dependency tree they derive is well-typed.

\begin{thm}[Soundness]
	For every goal configuration $c$ derived by LTF or LTL, the AM dependency tree described by $c$ is well-typed.
	\label{thm:sound}
\end{thm}
Here, "described by" means that we can read off the AM dependency tree from the set of edges $\pE_c$ and graph constants $\pG_{c}$. We do not need any additional assumptions to prove this theorem.

Before we can prove the theorem we first need the following lemma:

\begin{lemma}
	In every configuration $c$ derived by LTF or LTL, token $i$ has an $\app{\alpha}$ child
	if and only if $\alpha \in \pA_c(i)$.
	\label{lemma:aa}
\end{lemma}
\begin{proof}
	The $\trans{Apply}(\alpha, j)$ transitions in LTF and LTL always add an $\alpha$-source to $\pA_c(i)$ and simultaneously add an $\app{\alpha}$ edge. There are no other ways to add a source to $\pA_c(i)$ or to create an $\app{\alpha}$ edge.
\end{proof}

To prove the theorem, first observe that LTF and LTL only derive trees. 
Well-typedness then follows from applying the following lemma to the root of the tree in the goal configuration $c$:

\begin{lemma}
    Let $c$ be a goal configuration derived by LTF or LTL and $i$ be a token with $\pT_c(i)=\{\ttyp{t}\}$.
    Then the subtree rooted in $i$ is well-typed and has type $\ttyp{t}$.
\end{lemma}
\begin{proof}
By structural induction over the subtrees.
\paragraph{Base case}
 Since $i$ has no children, it has no $\app{}$ children in particular, making $\pA_c(i) = \emptyset$ by Lemma \ref{lemma:aa}. By definition of the goal configuration, $\pA_c(i) = \applyset(\tp{\pG_c(i)}, \ttyp{t})$. Combining this with $\pA_c(i) = \emptyset$, we deduce that $\tp{\pG_c(i)} = \ttyp{t}$ using the definition of the apply set.

\paragraph{Induction step}
Let $i$ be a node with \app{} children $a_1, \ldots,a_n$, attached with the edges $\app{\alpha_1}, \ldots, \app{\alpha_n}$, respectively. Let $i$ also have $\modify{}$ children $m_1, \ldots, m_k$, attached with the edges $\modify{\beta_1}, \ldots, \modify{\beta_k}$, respectively. Let $\lextyp{l} = \tp{\pG_{c}(i)}$ be the lexical type at $i$, and $\{\ttyp{t}\} = \pT_c(i)$.

By the definition of the apply set, $i$ reaches term type $\ttyp{t}$ from $\lextyp{l}$ if we can show for all \app{} children:
\begin{enumerate}[(i)]
    \item $i$ has an $\app{\alpha}$ child if and only if $\alpha \in \applyset(\lextyp{l}, \ttyp{t})$
    \item if $a$ is an $\app{\alpha}$ child of $i$, then it has the term type $\req{\lextyp{l}}{\alpha}$.
\end{enumerate}
(i)  follows from the goal condition $\pA_c(i) = \applyset(\lextyp{l}, \ttyp{t})$ and Lemma \ref{lemma:aa}. \\
(ii)  the only way the edge $\ledge{i}{\app{\alpha}}{a}$ can be created is by the $\trans{apply}(\alpha,a)$ transitions with $i$ on top of the stack. Both transition systems enforce $\pT_c(a) = \{\req{\lextyp{l}}{\alpha}\}$. Using the inductive hypothesis on $a$, it follows that $a$ evaluates to a graph of type $\req{\lextyp{l}}{\alpha}$. 

Although the \modify{} children of $i$ cannot alter the term type of $i$, they could make the subtree rooted in $i$ ill-typed. That is, for any \modify{\beta} child $m$ that evaluates to a graph of type $\ttyp{t'}$ by the inductive hypothesis, we have to show that $\ttyp{t}' - \beta \subseteq \lextyp{l} \wedge
\req{\ttyp{t}'}{\beta}=\emptytype$. The $\modify{\beta}$ edge was created by a $\trans{modify}(\beta,m)$ transition. The $\trans{modify}(\beta,m)$ transition (in case of LTF) or the next $\trans{Finish}$ transition (in case of LTL) resulted in a configuration $c'$, where the term types of $m$ were restricted in exactly that way: $\pT_{c'}(m) = \{ \tau \in \Omega |  \tau - \beta \subseteq \lextyp{l} \wedge
\req{\tau}{\alpha}=\emptytype \}$. In the derivation from $c'$ to $c$, a \trans{Choose} (LTF) or \trans{Finish} (LTL) transition must have been applied when $m$ was on top of the stack (because the $\modify{\beta}$ edge was created and $c$ is a goal configuration), which resulted in $\pT_{c}(m)=\{\ttyp{t'}\}$, where $\ttyp{t'} \in\pT_{c'}(m) =  \{ \tau \in \Omega |  \tau - \beta \subseteq \lextyp{l} \wedge
\req{\tau}{\alpha}=\emptytype \}$. This means that the well-typedness condition indeed also holds for  $\ttyp{t'}$.
\end{proof}

\subsection{Completeness}

\begin{thm}[Completeness]
	For every well-typed AM dependency tree $t$, there are valid sequences
	of LTF and LTL transitions that build exactly $t$.
	\label{thm:complete}
\end{thm}
We do not need any additional assumptions to prove this theorem.
The proof is constructive: for any well-typed AM dependency tree $t$, Algorithms \ref{alg:genLTF} and \ref{alg:genLTL} give transition sequences that, when prefixed with an appropriate \trans{Init} operation, generate $t$. We show this by showing the following lemma (for LTF):
\begin{lemma}
    Let $t$ be a well-typed AM dependency tree with term type $\ttyp{\tau}$ whose root is $r$ and let $c$ be a configuration derived by LTF with
    \begin{enumerate}[(i)]
    	\item $\ttyp{t}\in\pT_c(r)$,
    	\item $r$ is on top of $\pS_c$,
    	\item $W_c-O_c \geq |t|-1$, \ie $W_c-O_c $ is at least the number of nodes in $t$ without the root,
    	\item $i \notin Dom(\pG_c)$ for all nodes $i$ of $t$, and
    	\item $i \notin Dom(\pE_c)$ for all nodes $i \neq r$ of $t$
    \end{enumerate}
    Then $H_{LTF}(c,t)$ (Algorithm \ref{alg:genLTF}) constructs, with valid LTF transitions, a configuration $c'$ such that
    \begin{enumerate}[(a)]
    	\item $c'$ contains the edges of $t$,
    	\item $\pG_{c'}(i)=G_i$ where $G_i$ is the constant at $i$ in $t$,
    	\item $\pS_{c'}$ is the same as $\pS_{c}$ but without $r$ on top, \ie $\pS_{c} = \pS_{c'} | r$,
    	\item $ W_{c'} = W_c - (|t|-1)$, and 
    	\item for all $j$ that are \emph{not} nodes of $t$, none of $\pA, \pG, \pT, \pE$ changes, e.g.~$\pA_{c'}(j) = \pA_c(j)$ .
    \end{enumerate}
    \label{lemma:complete-ltf}
\end{lemma}
The lemma basically says that we can insert $t$ as a subtree into a configuration $c$ with LTF transitions. The conditions (i) and (ii) say that we have already put the root of $t$ on top of the stack and thus can now start to add the rest of $t$. Condition (iii) says that there are enough words left in the sentence to fit $t$ into $c$, where $-1$ comes from the fact that the root of $t$ is already on the stack and has an incoming edge. Conditions (iv) and (v) ensure that the part is still empty where we want to put the subtree.

Theorem \ref{thm:complete} for LTF then follows from applying the lemma to the whole tree $t$ and the configuration obtained after $\trans{Init}(t)$. This yields a configuration with empty stack, which is a goal configuration (see Lemma \ref{lemma:empty-stack-ltf}).
 
 Before we approach the proof of Lemma \ref{lemma:complete-ltf}, we need to show the following:
 
 \begin{lemma}
 	Let $c$ be a configuration derived by LTF. If for any token $i$, $i \notin \dom(\pG_c)$ then $i \notin \dom(\pA_c)$.
 	\label{lemma:pgpa-ltf}
 \end{lemma}
 \begin{proof}
 	We show its contraposition: If for any token $i$, $i \in \dom(\pA_c)$ then $i \in \dom(\pG_c)$. The \trans{Choose} transition defines $\pA_c$ for $i$, and defines $\pG_c$ for $i$ at the same time. There is no transition that can remove $i$ from $\dom(\pG_c)$.
 \end{proof}

\begin{proof}[Proof of Lemma \ref{lemma:complete-ltf}.]
	By structural induction over $t$.
		\paragraph{Base case}
		Let $i$ be on top of the stack in $\pS_{c}$. $t$ is a leaf with graph constant $G$, thus $W_c - O_c \geq |t|-1 = 0$. $H_{LTF}$ returns the sequence $\trans{Choose}(\tp{G}, G), \trans{Pop}$.
		It is easily seen that this sequence, if valid, yields a configuration $c'$ where $\pT_{c'}(i)=\{\tp{G}\}$, $\pG_{c'}(i) = G$ and $\pA_{c'}(i) = \applyset(\tp{G},\tp{G}) = \emptyset$. $c'$ also contains all edges of $t$ (there are none).
		
		 In order for $\trans{Choose}(\tp{G}, G)$ to be applicable, it must hold that $\tp{G} \in \pT_{c}(i)$ (holds by (i)), $i \notin \dom(\pG_{c})$ (holds by (iv)) and that $\tp{G} \in \possl(\tp{G}, \emptyset, W_c -O_c)$, which is equivalent to 
		 $$ |\applyset(\tp{G},\tp{G})| \leq W_c - O_c $$Since $\applyset(\tp{G},\tp{G}) = \emptyset$ and $W_c - O_c \geq 0$, this holds with equality. The transition $\trans{Choose}(\tp{G}, G)$ yields a configuration $c_1$, where $\pA_{c_1}(i) = \applyset(\tp{G},\tp{G}) = \emptyset$, so we can perform \trans{Pop}, which gives us the configuration $c'$. Since we have not drawn any edge $W_{c'} = W_c = W_c - (1 - 1) = W_c - (|t| - 1)$. 
		 Note that these transitions have not changed any $\pA, \pG, \pT, \pE$ for $j \neq i$.
		 
		 \paragraph{Induction step}
		 Let $i$ be on top of the stack in $\pS_{c}$ and let $i$ in $t$ have \app{} children $a_1, \ldots,a_n$, attached with the edges $\app{\alpha_1}, \ldots, \app{\alpha_n}$, respectively, where $n$ might be $0$.
		 Let $i$ in $t$ also have $\modify{}$ children $m_1, \ldots, m_k$, attached with the edges $\modify{\beta_1}, \ldots, \modify{\beta_k}$, respectively, where $k$ might be $0$ as well.
		 Let $G$ be the constant of $i$ in $t$, and $\ttyp{t}$ be its term type. By well-typedness of $t$ and the definition of the apply set, we have $\applyset(\tp{G}, \ttyp{t}) = \{\alpha_1, \ldots, \alpha_n\}$.
		 
		 $H_{LTF}(t, c)$ returns the sequence in Fig.~\ref{eq:monster-seq}, where $c_1$ is the configuration after $\trans{Choose}(\ttyp{t}, G), \trans{apply}(\alpha_1, a_1)$ etc.
         
         \begin{figure*}
		 \begin{nalign}
		 	 \bigedge{c}{c_0'}{\trans{Choose}(\ttyp{t}, G)} \bigedge{}{c_1}{$\trans{apply}(\alpha_1, a_1)$}
		 	 \bigedge{}{c_1'}{$H_{LTF}(a_1, c_1)$} \ldots \bigedge{c_{n-1}'}{c_n}{$\trans{apply}(\alpha_n, a_n)$} \bigedge{}{c_{n'}}{$H_{LTF}(a_n, c_n)$} \\
		 	\bigedge{c_{n}'}{c_{n+1}}{$\trans{modify}(\beta_1, m_1)$}
		 	\bigedge{}{c_{n+1}'}{$H_{LTF}(m_1, c_{n+1})$} \ldots 
		 	\bigedge{}{c_{n+k}'}{$H_{LTF}(m_k, c_{n+k})$}
		 	\bigedge{}{c'}{\trans{Pop}}
		 \end{nalign}
         \caption{Transition sequence returned by $H_{LTF}(t, c)$ in the induction step.}
         \label{eq:monster-seq}
         \end{figure*}

		  For now, let us assume that conditions (i)-(v) are fulfilled for $a_1, \ldots, a_n, m_1, \ldots, m_k$ and their respective configurations and that the sequence is valid. We will verify this at a later stage. 
		  
		  We can apply the inductive hypothesis for all children, which means that $c'$ contains the edges present in the subtrees $a_1, \ldots, a_n, m_1, \ldots, m_k$ and  for all nodes $j$ such that $j$ is a descendant of one of $a_1, \ldots, a_n, m_1, \ldots, m_k$, it holds that $\pG_{c'}(j)=G_j$ because $H_{LTF}$ applied to some child of $t$ will do the assignment and such an assignment can never be changed in LTF. Assuming the above transition sequence is valid, it is obvious that it also adds the edges from $i$ to $a_1, \ldots, a_n, m_1, \ldots, m_k$ with the correct labels (consequence (a)) and also makes the assignment $\pG_{c'}(i)=G_i$ using $\trans{Choose}(\ttyp{t}, G)$ (consequence (b)).
		  
		  Now we go over the transition sequence in Eq.~\ref{eq:monster-seq} and check that the transitions can be applied, the conditions (i)-(v) hold and what happens to the stack.
		  
		  First, in order for $\trans{Choose}(\tp{G}, \ttyp{t})$ to be applicable, it must hold that $\ttyp{t} \in \pT_{c}(i)$ (holds by (i)), $i \notin \dom(\pG_{c})$ (holds by (iv)) and that $\tp{G} \in \possl(\ttyp{t}, \emptyset, W_c -O_c)$, which is equivalent to 
		  $$ |\applyset(\tp{G},\ttyp{t})| \leq W_c - O_c $$
		  Since $\applyset(\tp{G},\ttyp{t}) = \{\alpha_1, \ldots, \alpha_n\}$ and $W_c - O_c \geq |t|-1 \geq |\{\alpha_1, \ldots, \alpha_n\}| = n$, this holds. This yields a configuration $c_0'$ where $\pT_{c_0'}(i) = \{\ttyp{t}\}$ and $\pA_{c_0'}(i) = \emptyset$.
		  
		  Next, we use the transition $\trans{apply}(\alpha_1,a_1)$. This is allowed because $\alpha_1 \in \applyset(\tp{G}, \ttyp{t})$ (see above), $\alpha_1 \notin \pA_{c_0'}(i)$ and $a_1 \notin \dom(\pE_{c_0'})$ (condition (v)). We get a new configuration $c_1$ where $\pA_{c_1}(i)=\{\alpha_1\}$, $\pT_{c_1}(a_1)=\{\req{\tp{G}}{\alpha_1}\}$ and $\pS_{c_1} = \pS_{c_0}|a_1$. We now justify why the inductive hypothesis can be used for $a_1$ and $c_1$:
		  
		  By well-typedness of $t$, we know that $\pT_{c_1}(a_1)=\{\req{\tp{G}}{\alpha_1}\} = \{\ttyp{t_{a_1}} \}$ where $\ttyp{t_{a_1}}$ is the term type of $a_1$ (condition (i)). From the step before, $a_1$ is on top of the stack in $\pS_{c_1}$ (condition (ii)). We use the fact that $j \notin Dom(\pG_c)$ for all nodes $j$ of $t$ and $j \notin Dom(\pE_c)$ for all nodes $j \neq i$ (our conditions (iv) and (v)) to justify that conditions (iv) and (v) are also met for $a_1$. What is left to verify is that $W_{c_1} - O_{c_1} \geq |a_1|-1$.
		  First, note that $W_{c_1} = W_c - 1$ because of the $\app{\alpha_1}$ edge. We can decompose $O_{c_1}$ as follows:
		  $$ O_{c_1} = O_c - O_c(i) + O_{c_1}(i) $$
		  because we have only changed $\pG_c$ and $\pA_c$ for $i$, not for any other token. $O_c(i) = 0$ by Lemma \ref{lemma:pgpa-ltf} and $i \notin \dom(\pG_c)$ (condition (iv)). We can also see that $O_{c_1}(i) = n-1$ by definition of $O(\cdot)$ and taking into account that we have drawn the  $\app{\alpha_1}$ edge and thus $\pA_{c_1} = \{\alpha_1\}$. This means that
		  $$ W_{c_1} - O_{c_1} = (W_c-1) - (O_c + n-1) = W_c - O_c - n $$
		  From condition (iii), we know that $W_c - O_c \geq |t|-1$. Since $t$ consists of node $i$ and at least $n$ children $a_j$ each of which has $|a_j|-1$ nodes, we have that
		  \begin{align*}
		  |t| \geq 1 + n + \sum_{j=1}^n (|a_j|-1)
		  \end{align*}
		  which is equivalent to
		  \begin{align}
		  |t|-1 \geq  n + \sum_{j=1}^n (|a_j|-1)
		  \label{eq:size-t}
		  \end{align}
		  Plugging this together, we get
		  $$ W_{c_1} - O_{c_1} = W_c - O_c - n \geq \sum_{j=1}^n (|a_j|-1) \geq |a_1|-1  $$ 
		  After $H_{LTF}(a_1,c_1)$ we get a configuration $c_1'$. We have just argued that the inductive hypothesis applies for $H_{LTF}(a_1,c_1)$, so we can use it and find that we are in a nearly identical situation as before $\trans{apply}(\alpha_1, a_1)$: The stack is $\pS_{c_1'} = \pS_{c_1} | a_1 = \pS_{c}$. That is, in $\pS_{c_1'}$ the top of the stack is $i$ again. What has changed is $W_{c_1'} -  O_{c_1'}$ and of course $\pA_{c_1'} = \{\alpha_1\}$, which was empty before. We can now apply $\trans{apply}(\alpha_2, a_2)$ and continue.
		  
		  Let us consider the general case for $H_{LTF}(a_l, c_l)$ with $1 \leq l \leq n$ where we are in $c_l$ arriving from $\trans{apply}(\alpha_l, a_l)$. At this point, we know
		  \begin{enumerate}[(i)]
		  	\item $\pT_{c_l} = \{\ttypO{\tau_{a_{l}}}\}$ where $\ttypO{\tau_{a_{l}}}$ is the term type of $a_l$ (by $\trans{apply}$ before)
		  	\item $i$ is on top of the stack (inductive hypothesis for $l' < l$)
		  \end{enumerate}
		  In effect, conditions (i) and (ii) for the inductive hypothesis for $H_{LTF}(a_l, c_l)$ are met. Conditions (iv) and (v) for $a_l$ are fulfilled by our assumptions (iv) and (v) because $a_l$ is a subtree of $i$. What remains to be checked is $W_{c_l} - O_{c_l} \geq |a_l|-1$. We can calculate $W_{c_l} = W_c - l - \sum_{j=1}^{l-1}(|a_j|-1)$, where the summation over $j$ comes from the inductive hypothesis for the children $j < l$ and $-l$ comes from the \trans{apply} transitions we have performed. $O_{c_l}$ is simply $O_{c_l} = O_c +n -l$ because the $\trans{Choose}$ transition resulted in $O_{c_0'}  = O_c + n$ and we have drawn $l$ $\app{}$ edges already.
		  Plugging this together, we get
		  \begin{align*}
		   W_{c_l} - O_{c_l} &= W_c -l - \sum_{j=1}^{l-1}(|a_j|-1) - (O_c +n -l) \\
		    		& \geq (|t|-1) -n -\sum_{j=1}^{l-1}(|a_j|-1) \\
		    		& \geq \sum_{j=l}^{n} (|a_j|-1) \geq |a_l| -1
		   \end{align*}
		  where the first step replaces $W_c-O_c$ by $|t|-1$ (assumption (iii)) and the second step replaces $(|t|-1)$ using Eq.~\ref{eq:size-t}.
		  
		  A similar line of reasoning can be used to justify the use of the inductive hypothesis for $H_{LTF}(m_1,c_{n+1}), \ldots, H_{LTF}(m_1,c_{n+k})$.
		  
		  Note that by applying the inductive hypothesis to all children, we know that $i$ is always on top of the stack after $H_{LTF}$ was applied. This justifies the final $\trans{Pop}$ transition, because at that point $\pA_{c_{n+k}'} = \applyset(\tp{G},\ttyp{t})$. Consequence (c) follows from this $\trans{Pop}$.

 		  We did not change any of $\pE, \pA,\pT, \pG$ outside of our subtree $i$ (consequence (e)). This follows from the inductive hypotheses of the children and the fact that $i$ was always on top of the stack when we performed any transition.
 		  
		  
		  If we want to determine $W_{c'}$, we note that we have drawn $n+k$ edges and for each child $ch \in a_1, \ldots, a_n, m_1, \ldots m_k$, we know by the inductive hypothesis that this has drawn $|ch|-1$ edges. In total, we have
		  \begin{align*}
		  W_{c'} & = W_c - \left[\sum_{j=1}^n (|a_j|-1) + \sum_{j=1}^{k} (|m_j|-1) \right]  \\
            & \quad - (n+k) \\
		   & = W_c - \left[\sum_{j=1}^n |a_j| + \sum_{j=1}^{k} |m_j| - (n+k) \right] \\
           & \quad - (n+k) \\
		   & = W_c - (|t|-1)
		  \end{align*}
		  where the last step makes use of the fact that $|t| = 1 + \sum_{j=1}^n |a_j| + \sum_{j=1}^{k} |m_j|$.
		  
		  
\end{proof}

For LTL, the same principle applies with a near identical lemma which only also asks that for the root $r$ of $t$, $\pA_c(r)=\emptyset$. The procedure to construct the transition sequence is shown in Algorithm \ref{alg:genLTL}.


\begin{algorithm}[t]
	\caption{Generate LTF transitions for AM dependency tree}
	\begin{algorithmic}[1] 
		\Function{$H_{LTF}$}{$c, t$}
		\State Let $t$ have graph constant $G$
		\State and term type $\ttyp{t}$
		\State $c \gets \trans{Choose}(\ttyp{t}, G)(c)$
		\For{$\app{\alpha}$ child $a$ of t}
		    \State $c \gets \trans{apply}(\alpha,a)(c)$
		    \State $c \gets H_{LTF}(c,a)$
		\EndFor
		\For{$\modify{\beta}$ child $m$ of $t$}
		    \State $c \gets \trans{modify}(\beta,m)(c)$
		    \State $c \gets H_{LTF}(c,m)$
		\EndFor
		\State $c \gets \trans{Pop}(c)$
		\State \textbf{return} $c$
		\EndFunction
	\end{algorithmic}
	\label{alg:genLTF}
\end{algorithm}

\begin{algorithm}[t]
	\caption{Generate LTL transitions for AM dependency tree}
	\begin{algorithmic}[1] 
		\Function{$H_{LTL}$}{$c, t$}
		\State Let $t$ have graph constant $G$
		\For{$\app{\alpha}$ child $a$ of t}
		    \State $c \gets \trans{apply}(\alpha,a)(c)$
		\EndFor
		\For{$\modify{\beta}$ child $m$ of $t$}
		    \State $c \gets \trans{modify}(\beta,m)(c)$
		\EndFor
		\State $c \gets \trans{Finish}(G)(c)$
		\State Let $t_1, \ldots t_n$ be the children of $t$
		\State on the stack in $c$
		\For{$ i \in 1, \ldots, n$}
		    \State $c \gets H_{LTL}(c,t_i)$
		\EndFor
		\State \textbf{return} $c$
		\EndFunction
	\end{algorithmic}
	\label{alg:genLTL}
\end{algorithm}

\subsection{No dead ends}

\input{no-dead-ends}

%% file: no-dead-ends.tex
For both LTF and LTL, the following theorem guarantees that we can always get a complete analysis for a sentence:
\begin{thm}[No dead ends]
	If $c$ is a configuration derived by LTF or LTL then there is a valid sequence of transitions that brings $c$ to a goal configuration $c'$.
	\label{thm:no-dead-ends}
\end{thm}
Together with the soundness theorem (Theorem \ref{thm:sound}) that every goal configuration corresponds to well-typed AM dependency tree, this means that we can always finish a derivation to get a well-typed AM dependency tree, no matter what the sentence is or how the transitions are scored. The proof of Theorem \ref{thm:no-dead-ends} is constructive both for LTF and LTL and is given below. In both cases, we proof a lemma first that there are always "enough" words left.

Theorem \ref{thm:no-dead-ends} only holds if we make a few assumptions that are mild in practice.
Recall that we assumed that we are given a set of graph constants $\Cons$ that can draw source names from a set $\Sources$, a set of types $\Omega$ and a set of edge labels $\edgeLabels$. We now make very explicit the following assumptions about their relationships:

\begin{assumption}
	For all types $\lextyp{l} \in \Omega$, there is a constant $G \in C$ with type $\tp{G} = \lextyp{l}$.
	\label{assumption:constants}
\end{assumption}
\begin{assumption}
	For all types $\lextyp{l} \in \Omega$ and all source names $\alpha \in \Sources$, if $\req{\lextyp{l}}{\alpha}$ is defined then $ \req{\lextyp{l}}{\alpha} \in \Omega$.
	\label{assumption:requests}
\end{assumption}
\begin{assumption}
	If $\modify{\alpha} \in Lab$ then $[\alpha] \in \Omega$.
	\label{assumption:mod}
\end{assumption}
\begin{assumption}
	For all source names $\alpha \in \Sources$, $\app{\alpha} \in Lab$.
	\label{assumption:app}
\end{assumption}
\begin{assumption}
	There are \textit{no} constraints imposed on which graph constants can be assigned to a particular word.
	\label{assumption:constants-words}
\end{assumption}

The assumptions made are almost perfectly
met in practice, see the main paper.

In the proof of Theorem \ref{thm:no-dead-ends} we want to use the fact $\emptytype \in \Omega$; this follows from the assumptions above:
\begin{lemma}
	The empty type $\emptytype \in \Omega$.
	\label{lemma:emptytype}
\end{lemma}
\begin{proof}
	Assumption \ref{assumption:requests} says that for all types $\lextyp{l} \in \Omega$ and all sources $\alpha \in \Sources$, the type $\req{\lextyp{l}}{\alpha}$ (if defined) is also a member of $\Omega$. Since types are formally DAGs, each type $\tau$ is either empty (that is: $\emptytype$) or has a node $n$ without outgoing edges. In the latter case, $\req{\tau}{n} = \emptytype$.
\end{proof}

\subsubsection{LTF}


We prove a lemma that there are always at most as many sources that we have still to fill as there are words without incoming edges. 

\begin{lemma}
	For all configurations derived with LTF, $O_c \leq W_c$.
	\label{lemma:ocwc}
\end{lemma}
\begin{proof}
	By structural induction over the derivation.
	\paragraph{Base case}
		  The initial state $c$ does not define $\pA$ for any token, thus $O_c(i)=0$ for all $i$. The number of words without incoming edges in configuration $c$ is $W_c \geq 1$. Therefore, $\sum_{i} O_c(i) = O_c \leq W_c$. 
	\paragraph{Induction step}
	Inductive hypothesis: $O_c \leq W_c$\\
		Goal: $O_{c'} \leq W_{c'}$ where $c'$ derives in one step from $c$. \\
		The derivation step from $c$ to $c'$ is one of:
		\begin{description}
			\item[$\trans{Init}(i)$] After \trans{init}, $\pA_{c'}$ is not defined for any $i$, thus $O_c = 0$.
			\item[\trans{Pop}] This transition only changes the stack, which does not affect $O$, so $O_{c'}(i) = O_c(i)$ for all $i$ and $W_{c'} = W_c$. The inductive hypothesis applies.
			
			
			\item[$\trans{Choose}(\ttyp{t},G)$] Let $i$ be the active node. No edge was created, thus $W_{c'} = W_c$.
			For all $j \neq i$, $O_{c'}(j) = O_c(j)$. We can thus write $O_{c'}$ as
			\begin{align*}
			O_{c'} = O_c -O_c(i) + O_{c'}(i)
			\end{align*}
			Since $\trans{Choose}(\ttyp{t},G)$ was applicable in $c$, we know that $i \notin \dom(\pG_{c})$. By Lemma \ref{lemma:pgpa-ltf} and by definition of $\possl$, we have that $O_c(i)=0$, so 
			\begin{align}
			O_{c'} = O_c + O_{c'}(i)
			\label{eq:oc-decomp-ltf}
			\end{align}
			
			We now look into the value of $O_{c'}(i)$.
			Since $\trans{Choose}$ was applied, we know that $\pG_{c'}(i)=G$, $\pA_{c'}(i) = \emptyset$ and that $\tp{G} \in \possl(\ttyp{t}, \emptyset, W_c-O_c)$, which simplifies to $|\applyset(\tp{G},\ttyp{t}) | \leq W_c-O_c $.
			From this follows that $O_{c'}(i) = \min_{\lextyp{l'} \in \{\tp{G}\}, \ttyp{t'} \in \pT_{c'}(i)} |\applyset(\lextyp{l'}, \ttyp{t'}) - \pA_{c'}(i)| \leq W_c-O_c$.
			Substituting this for $O_{c'}(i)$ in Eq.~\ref{eq:oc-decomp-ltf}, we get \begin{align*}
				 O_{c'} &= O_c + O_{c'}(i) \\
				 &  \leq O_c + W_c-O_c  = W_c = W_{c'}
				 \end{align*}
			\item[$\trans{Apply}(\alpha, j)$] Let $i$ be the active node. Since an edge to $j$ was created in the transition, $W_{c'}+1 = W_c$.
			We decompose $O_{c'}$ again:
			$$O_{c'} = O_c -O_c(i) + O_{c'}(i) $$
			Since $\trans{Apply}$ could be performed, we know that $\pT_c$ and $\pG_c$ are defined for $i$ and let us denote them $\pT_c(i) = \{\ttyp{t}\}$ and $\pG_c(i) = G$. Thus, $O_c(i) = | \applyset(\tp{G}, \ttyp{t}) - \pA_c(i)|$. Since the precondition of \trans{apply} said that $\alpha \notin \pA_c(i)$ and \trans{Apply} has the effect that $\pA_{c'}(i) = \pA_c(i) \cup \{\alpha\}$, we know that $O_{c'}(i) = | \applyset(\tp{G}, \ttyp{t}) - (\pA_c(i) \cup \{\alpha)\} | < O_c(i)$. This means that $O_{c'} < O_c$. Using the inductive hypothesis $O_c \leq W_c$ and $W_{c'}+1 = W_c$, we get
			\begin{align*}
				O_{c'} < O_c \leq W_{c'}+1
			\end{align*}
			which means that $O_{c'} \leq W_{c'}$.
			
			\item[$\trans{Modify}(\beta, j)$] Let $i$ be the active node. In Section \ref{sec:ltf}, we made the restriction that $\trans{Modify}$ is only applicable if \begin{align}
			W_c - O_c \geq 1
			\label{eq:modify}
			\end{align}
			The transition created an edge, which means that $W_{c'} = W_c-1$. $O_{c'}$ depends on $\pG_{c'}, \pA_{c'}$ and $\pT_{c'}$. The only thing that changed from $c$ to $c'$ is that $\pT_{c'}$ is now defined for $j$. However, $\pA_{c'}$ is still not defined for $j$, so $O_{c'}(j) = O_{c}(j) = 0$. This means $O_{c'} = O_c$. Substituting those into Eq.~\ref{eq:modify} and re-arranging, we get $O_{c'} \leq W_{c'}$.
		\end{description}
		
\end{proof}


We now show that there are no dead ends by showing that for any configuration $c$ derived by LTF, we can construct a valid sequence of transitions such that the stack becomes empty. By Lemma \ref{lemma:empty-stack-ltf} this means that $c$ is a goal configuration. We empty the stack by repeatedly applying Algorithm \ref{alg:completeLTF}.

In line \ref{line:19}, we compute the sources that we still have to fill in order to pop $i$ off the stack. We assume an arbitrary order and $o_j$ refers to one particular source in $o$. The symbol $\concat$ denotes concatenation.

\begin{lemma}
	For any configuration $c$, $C_{LTF}(c)$ (Algorithm \ref{alg:completeLTF}) generates a valid sequence $s$ of LTF transitions such that ($|\pS_{c'}| < |\pS_c|$ or $|\pS_{c'}| = 0$) and there is a token $i$ for which $\pG_{c'}(i)$ is defined, where $c'$ is the configuration obtained by applying $s$ to $c$.
	\label{lemma:make-stack-empty-ltf}
\end{lemma}

\begin{proof}
	First, we show that $\pG_{c'}$ is defined for some $i$ in $c'$.
	We make a case distinction based on in which line the algorithm returns. If it returns in lines \ref{line:exit-1}, \ref{line:exit-2} or \ref{line:exit-3}, it is obvious that $\pG_{c'}$ is defined for some $i$. If it returns in line \ref{line:exit-4} then $o$ is non-empty because $O_c(i)>0$. If $o$ is non-empty, we use a \trans{Choose} transition in the for-loop. The remaining case is returning in line \ref{line:exit-argument}. Note that the stack is empty but it is not the initial configuration (otherwise, we would have returned in line \ref{line:exit-1}), so an \trans{Init} transition must have been applied, which pushes a token to the stack. Since the stack is now empty in $c'$, a \trans{Pop} transition must have been applied, which is only applicable if $\pG$ is defined for the item on top of the stack. 
	Consequently, $\pG_{c'}$ is defined for some $i$. 
	
	Further, note that every path through Algorithm \ref{alg:completeLTF} either reduces the size of the stack (one more $\trans{Pop}$ transition than tokens pushed to the stack by $\trans{Apply}$) or keeps it effectively empty. 
	
		$C_{LTF}$ is constructed in a way that the transition sequence is valid.
	 However, there are a few critical points:
	\begin{itemize}
		\item In line \ref{line:empty-type}, we assume the existence of a graph constant $G \in \Cons$ with $\tp{G} = \emptytype$. This follows from Lemma \ref{lemma:emptytype} and Assumption \ref{assumption:constants}.
		
		\item In line \ref{line:15}, it is assumed that there exists a graph constant $G \in \Cons$ with $\tp{G} \in \pT_c(i)$. This graph constant always exists because either $\pT_c(i)$ is a request (if $i$ has an incoming $\app{}$ edge) and thus by Assumptions \ref{assumption:constants} and \ref{assumption:app} there is a graph constant $G \in C$, or $\pT_c(i)$ is a set of types resulting from a \trans{Modify} transition. Here, the existence of a suitable graph constant $G$ with type $\tp{G} \in \pT_c(i)$ follows from Assumptions \ref{assumption:constants} and \ref{assumption:mod}. Assumption \ref{assumption:constants-words} makes explicit that there are no further constraints on how we choose $G$.
		\item In line \ref{line:22}, it is assumed that there exist $|o|$ tokens without incoming edges. This is true because $|o| = O_c(i) \leq \sum_{j} O_c(j) = O_c$ and by Lemma \ref{lemma:ocwc}, it follows that $|o| \leq W_c$, showing that there are indeed enough tokens without incoming edges.
		\item In line \ref{line:24}, it is assumed that $\app{a_i} \in \edgeLabels$ for some source $o_j$; this is guaranteed by Assumption \ref{assumption:app}.
	\end{itemize}

\end{proof}

In summary, we can turn any configuration $c$ derived by LTF into a goal configuration by repeatedly applying $C_{LTF}$ to it until the (finite) stack is empty. By Lemma \ref{lemma:empty-stack-ltf}, this is a goal configuration.

\begin{algorithm}[t]
	\caption{Complete LTF sequence}
	\begin{algorithmic}[1] 
		\Function{$C_{LTF}$}{$c$}
		\If{$c = \langle \emptyset, \emptyset, \emptyset, \emptyset, \emptyset
			\rangle$}
		\State Let $G \in \Cons$ with $\tp{G} =\emptytype$.
		\label{line:empty-type}
		\State \textbf{return} $\trans{Init}(1), \trans{choose}(\emptytype,G), \trans{Pop}$ 
		\label{line:exit-1}
		\EndIf
		\If{$\pS_c = []$}
		\Return $[]$
		\label{line:exit-argument}
		\EndIf
		\State Let $i$ be top of $\pS_c$.
		\If{$O_c(i) = 0$}
		\If{$i \in Dom(\pG_c)$}
		\State \textbf{return} $\trans{Pop}$
		\label{line:exit-2}
		\Else
		\State Let $G \in \Cons, \tp{G} \in \pT_c(i)$.
		\label{line:15}
		\State \textbf{return} $\trans{choose}(\tp{G},G), \trans{Pop}$
		\label{line:exit-3}
		\EndIf
		\EndIf
		\State Let $o = \applyset(\pG_c(i), \tau) - \pA_c(i)$
		\label{line:19}
		\State where $\pT_c(i) = \{\tau\}$
		\State Let $\rho_j = \req{\tp{\pG_c(i)}}{o_j}$
		\State Let $a_1, \ldots, a_{|o|}$ be tokens without heads
		\label{line:22}
		\State $s = []$
		\For{$a_j \in a_1, \ldots, a_{|o|}$}
		\State Let $G$ be a constant of type $\rho_j$
		\State $s = s \concat \trans{apply}(o_j, a_j), \trans{Choose}(\rho_j,G), \trans{Pop}$
		\label{line:24}
		\EndFor
		\State \textbf{return} $s \concat \trans{Pop}$
		\label{line:exit-4}
		
		\EndFunction
	\end{algorithmic}
	\label{alg:completeLTF}
\end{algorithm}

\subsubsection{LTL}

The proof works similarly. We first prove a similar lemma that if $i$ is the active node, $O_c(i) \leq W_c$ and then construct a function $C_{LTL}$ (see Algorithm \ref{alg:completeLTL}) that produces a valid sequence of transitions that we repeatedly apply to reach a goal configuration.

\begin{lemma}
	Let $c$ be a configuration derived by LTL.
	If $i$ is the active node in $c$, then $O_c(i) \leq W_c$.
	\label{lemma:ocwc-ltl}
\end{lemma}
\begin{proof}
	By structural induction over the derivation.
	\paragraph{Base case}
		 In the initial state, the stack $\pS_c$ is empty, making the antecedent of the implication false for all $i$ and thus the implication true.
	\paragraph{Induction step}
		 Inductive Hypothesis:	If $i$ is the active node in $c$, then $O_c(i) \leq W_c$. \\
		Goal: If $i$ is the active node in $c'$, then $O_{c'}(i) \leq W_{c'}$ where $c'$ derives in one step from $c$. \\
		The applied transition is one of:
		\begin{description}
				\item[$\trans{Init}(i)$] The previous configuration $c$ must be the initial configuration. Now $i$ is the active node in $c'$ and $\pA_{c'}(i) = \emptyset$ and $\pT_{c'}(i) = \{\emptytype\}$ and $\pG_{c'}$ is not defined for $i$. Then $O_{c'}(i) = \min_{\lextyp{l} \in \Omega} |\applyset(\lextyp{l}, \emptytype) -  \pA_{c'}(i)| $.
				 Note that the empty type $\emptytype \in \Omega$ by lemma \ref{lemma:emptytype} and that $\applyset(\emptytype, \emptytype) = \emptyset$. Choosing $\lextyp{\lambda} = \emptytype$, we get $O_{c'}(i) = 0$.  $\trans{Init}(i)$ created an edge into $i$, so $W_{c'} = W_c -1$. Since a sentence consists of at least one word ($W_c \geq 1$), we have $O_{c'}(i) = 0 \leq W_{c'} $.
				 
				 \item[$\trans{Apply}(\alpha, j)$] Let $i$ be the active node in $c$. Then, by construction of $\trans{Apply}(\alpha, j)$ it remains the active node in $c'$. After the transition, $\pT_{c'}(i) = \pT_c(i)$, $\pA_{c'}(i) = \pA_{c}(i) \cup \{\alpha\}$. Thus, $O_{c'}(i)$ can be written as follows:
				 $$O_{c'}(i) = \min_{\lextyp{l'} \in \Omega, \ttyp{t'} \in \pT_{c}(i)} |\applyset(\lextyp{l'}, \ttyp{t'}) - (\pA_{c}(i) \cup \{\alpha\})| $$
				 
				  Since $\trans{Apply}(\alpha, j)$ was applicable, the pre-conditions must be fulfilled, i.e.\ 
				 \begin{align*}
				 	\exists \lextyp{l} \in \Omega. & \exists \ttyp{t} \in \pT_c(i). \\ 
                    & \lextyp{l} \in \possl(\ttyp{t}, \pA_{c}(i) \cup \{\alpha\}, W_c-1)
				 \end{align*}
				 Expanding the definition of $\possl$ we get:
				 \begin{align*}
                \pA_{c}(i) \cup & \{\alpha\} \subseteq \applyset(\lextyp{l}, \ttyp{t})  \wedge \\ &  |\applyset(\lextyp{l}, \ttyp{t}) - (\pA_{c}(i) \cup \{\alpha\})| \leq W_c-1
				 \end{align*}
				 for some $\lextyp{l}\in \Omega$ and $\ttyp{t} \in \pT_c(i)$.
				 If we now choose $\lextyp{l'} = \lextyp{l}$ and $\ttyp{t'} = \ttyp{t}$ in $O_{c'}(i)$, we get
				 $$O_{c'}(i) \leq  |\applyset(\lextyp{l}, \ttyp{t}) - (\pA_{c}(i) \cup \{\alpha\})| \leq W_c-1 $$
				 Since $W_{c'} = W_c - 1$, it holds that $O_{c'}(i) \leq W_{c'}$.
				 
				 \item[$\trans{Modify}(\beta, j)$] Let $i$ be the active node. It also remains the active node in $c'$. The transition consumes a word, that is $W_{c'} = W_c-1$. However, it can only be applied if $W_c - O_c \geq 1$. Since $O_c$ is obtained by summing over all tokens, $O_c(i) \leq O_c$. We get:
				 $$ O_c(i) \leq O_c \leq W_c - 1 = W_{c'}. $$
				 Finally, $O_{c'}(i) = O_c(i)$ because none of $\pA, \pG, \pT$ changed for $i$ during the $\trans{Modify}(\beta, j)$ transition.
				 
				 \item[$\trans{Finish}(G)$] Let $i$ be active node \textit{after} the transition, that is, in $c'$. The \trans{Finish} transition presupposes that $i$ has an incoming edge.
				 We distinguish two cases based on the label:
				 \begin{itemize}
				 	\item $i$ has an incoming $\app{\alpha}$ edge. Then we have that $\pT_{c'}(i)=\{\req{\tp{G}}{\alpha}\}$ and $\pG_{c'}$ undefined for $i$.
				 	Then $O_{c'}(i) = \min_{\lextyp{l} \in \Omega} |\applyset(\lextyp{l}, \req{\tp{G}}{\alpha}) |$.
				 	By Assumption \ref{assumption:requests}, $\req{\tp{G}}{\alpha} \in \Omega$ and by definition of the apply set $\applyset(\lextyp{l}, \lextyp{l}) = \emptyset$ for all types $\lextyp{l}$, so in particular also for $\req{\tp{G}}{\alpha}$, which makes $O_{c'}(i)=0$.
				 	\item $i$ has an incoming $\modify{\beta}$ edge. By Assumption \ref{assumption:mod}, we know that $\type{\beta} \in \Omega$, for which $\type{\beta} \in \pT_{c'}(i)$ holds by construction of $\trans{Finish}(G)$. Expanding the definition of $O_{c'}(i)$, we get: $O_{c'}(i) = \min_{\lextyp{l} \in \Omega, \ttyp{t'} \in \pT_{c'}(i)} |\applyset(\lextyp{l},\ttyp{t'}) |$. By choosing $\lextyp{\lambda} = \type{\beta} = \ttyp{t'}$, we get $O_{c'}(i) = 0$.

				 \end{itemize}
			 	Since $O_{c'}(i) = 0$, it also holds that $O_{c'}(i) \leq W_{c} = W_{c'} $.

		\end{description}
		
\end{proof}

\begin{algorithm}[t]
	\caption{Complete LTL sequence}
	\begin{algorithmic}[1] 
		\Function{$C_{LTL}$}{$c$}
		\If{$c = \langle \emptyset, \emptyset, \emptyset, \emptyset, \emptyset
			\rangle$}
		\State Let $G \in \Cons$ with $\tp{G} = \emptytype$
		\label{line:empty-type-ltl}
		\State \textbf{return} $\trans{Init}(1), \trans{Finish}(G)$ 
		\EndIf
		\If{$\pS_c = []$}
		\label{line:no-graph-constant-ltl}
		\Return $[]$
		\EndIf
		\State Let $i$ be top of $\pS_c$.
		\State Let $\lextyp{l},\ttyp{t}$ be the minimizers of $O_c(i) = \min_{\lextyp{l} \in \Omega, \ttyp{t} \in \pT_c(i)}
		|\applyset(\lextyp{l}, \ttyp{t}) - \pA_c(i)|$
		\label{line:ltexist}
		\If{$O_c(i) = 0$}
		\State Let $G \in \Cons$ with $\tp{G} = \lextyp{l}$
		\label{line:12ltl}
		\State \textbf{return} $\trans{finish}(G)$
		\label{line:return-non-empty-1-ltl}
		\EndIf
		\State Let $o = \applyset(\lextyp{l}, \ttyp{t}) - \pA_c(i)$
		\label{line:15ltl}
		\State Let $\rho_i = \req{\tp{\pG_c(i)}}{o_i}$
		\State Let $a_1, \ldots, a_{|o|}$ be tokens without heads
		\label{line:17ltl}
		\State $s = []$
		\For{$a_j \in a_1, \ldots, a_{|o|}$}
		\State $s = s \concat \trans{apply}(o_j, a_j)$
		\label{line:20ltl}
		\EndFor
		\State $s = s \concat \trans{Finish}(G)$ where $\tp{G} = \lextyp{l}$
		\label{line:other-ltl}
		\State \textbf{return} $s$
		\label{line:return-non-empty-2-ltl}
		
		\EndFunction
	\end{algorithmic}
	\label{alg:completeLTL}
\end{algorithm}

\begin{lemma}
	For a sentence with $n$ words, a valid LTL transition sequence can contain at most $n$ $\trans{Finish}$ transitions.
	\label{lemma:finish-limit}
\end{lemma}
\begin{proof}
	By contradiction. Assume there is a valid transition sequence $s$ that contains $m>n$ $\trans{Finish}$ transitions. \\
	Since $\trans{Finish}$ can only be applied when there is some token on the stack and there are more $\trans{Finish}$ transitions than there are tokens, $\trans{Finish}$ must have been applied twice with the same active node. Since $\trans{Finish}$ removes the active node from the stack, $i$ must have been pushed twice. This means that $i$ has two incoming edges. When the second incoming edge was drawn into $i$ the condition $i \notin \dom(\pE)$ was violated, which contradicts the assumption that the transition sequence $s$ is valid.

\end{proof}

\begin{lemma}
	Let $c$ be a configuration derived by an LTL transition sequence $s$ that contains $j$ $\trans{Finish}$ transitions. Then $C_{LTL}(c)$ (Algorithm \ref{alg:completeLTL}) generates a valid sequence $s'$ of LTL transitions that leads to a goal configuration $c'$ or $s \concat s'$ contains $j+1$  $\trans{Finish}$ transitions.
	\label{lemma:empty-or-finish} 
\end{lemma}

\begin{proof}
	We first show the main claim and then verify that the generated transition sequence $s'$ is valid.
	\noindent
	We make a case distinction on the content of the stack in $c'$.
	\begin{description}
		\item[$\pS_{c'}$ is empty] We show that $c'$ is a goal configuration. In order to apply Lemma \ref{lemma:empty-stack-ltl}, we have to show that $\pG_{c'}$ is defined for some token $i$. 	There is only one path through Algorithm \ref{alg:completeLTL} that does not assign a graph constant to a token (line \ref{line:no-graph-constant-ltl}). Returning in line \ref{line:no-graph-constant-ltl} means that the stack is empty but the state is not the initial state -- so something has been removed from the stack already with a \trans{Finish} transition. Consequently, $\pG$ is defined for some $i$.
		\item[$\pS_{c'}$ is not empty] Since the stack is not empty, this means the algorithm returns in line \ref{line:return-non-empty-1-ltl} or in line \ref{line:return-non-empty-2-ltl}. Clearly, the transition sequence that the algorithm returns contains a $\trans{Finish}$ transition. Together with the $j$ $\trans{Finish}$ transitions that have been performed up to the configuration $c$, this makes $j+1$ $\trans{Finish}$ transitions.
	\end{description}
	Algorithm \ref{alg:completeLTL} is constructed such that it only produces valid transition sequences. However, there are a few critical points:
	\begin{itemize}
		\item Line \ref{line:empty-type-ltl} assumes the existence of a graph constant $G \in \Cons$ with $\tp{G} = \emptytype$. This follows from Lemma \ref{lemma:emptytype} and Assumption \ref{assumption:constants}. Assumption \ref{assumption:constants-words} explicitly allows us to assign $G$ to any token.
		\item Line \ref{line:ltexist} assumes that $K_c(i) = \Omega$ and that $i \in \dom(\pA_{c})$ and $i \in \dom(\pT_{c})$. This is true because $i$ is on top of the stack. $\pA$ and $\pT$ are always defined for the active node in LTL. $\pG$ is never defined for the active node in LTL.
		\item Lines \ref{line:12ltl} and \ref{line:other-ltl} assume the existence of a graph constant $G \in \Cons$ of type $\tp{G} = \lextyp{l} \in \Omega$, which is guaranteed by Assumption \ref{assumption:constants}. Assumption \ref{assumption:constants-words} explicitly allows us to assign $G$ to any token.
		\item Line \ref{line:17ltl} assumes that there are at least $|o|$ tokens without incoming edges ($W_c \geq |o|$). This is indeed the case, because $|o| = O_c(i)$ and $O_c(i) \leq W_c$ by Lemma \ref{lemma:ocwc-ltl}.
		\item Line \ref{line:20ltl} assumes that $\app{o_j} \in \edgeLabels$. This is guaranteed by Assumption \ref{assumption:app}.
	\end{itemize}
\end{proof}
We can construct the transition sequence for which Theorem \ref{thm:no-dead-ends} asks by repeatedly applying $C_{LTL}$ to a given configuration $c$. Lemma \ref{lemma:empty-or-finish} shows that applying $C_{LTL}$ to a configuration results either in a goal configuration or increases the number of $\trans{Finish}$ transitions by one. Lemma \ref{lemma:finish-limit} tells us that there is an upper bound on how many times we can increase the number of $\trans{Finish}$ transitions in a valid transition sequence. Since $C_{LTL}$ returns only valid transition sequences, this means that we reach a goal configuration by finitely many applications of $C_{LTL}$.
